%% file: main.tex
\documentclass{egpubl}
\usepackage{pg2025s}

\WsConferencePaper

\usepackage[T1]{fontenc}
\usepackage{dfadobe}  

\usepackage{cite}  %
\BibtexOrBiblatex
\electronicVersion
\PrintedOrElectronic
\ifpdf \usepackage[pdftex]{graphicx} \pdfcompresslevel=9
\else \usepackage[dvips]{graphicx} \fi

\usepackage{egweblnk}

\title[Stylistic Attribute Control in LDMs]%
      {Stylistic Attribute Control in Latent Diffusion Models}

\author[M. Reimann et al.]
{\parbox{\textwidth}{\centering Max Reimann
        and Benito Buchheim
        and Jürgen Döllner
        }
        \\
{\parbox{\textwidth}{\centering Hasso-Plattner-Institute, University of Potsdam, Germany\\
       }
    }
}

\usepackage[nolist]{acronym}

\usepackage{amsmath}
\usepackage{amssymb}
\usepackage{booktabs}
\usepackage{wrapfig}
\usepackage[export]{adjustbox} %
\usepackage{caption}
\usepackage{subcaption}
\usepackage{placeins}

\usepackage[capitalize]{cleveref}
\crefname{section}{Sec.}{Secs.}
\Crefname{section}{Section}{Sections}
\Crefname{table}{Table}{Tables}
\crefname{table}{Tab.}{Tabs.}
\crefname{figure}{Fig.}{Figs.}

\input{acronym}

\begin{document}

\teaser{
\centering
\begin{minipage}{\textwidth}
    \centering
    \begin{minipage}{0.1625\linewidth}
        \includegraphics[width=\linewidth, height=2.8cm]{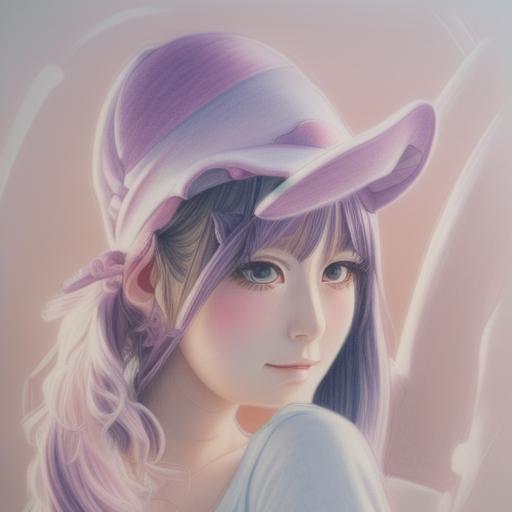}
        \vspace{-1em} %
        \captionsetup{type=figure, skip=0pt}\caption*{SD unedited output}
    \end{minipage}
    \hfill
    \begin{minipage}{0.1625\linewidth}
        \includegraphics[width=\linewidth, height=2.8cm]{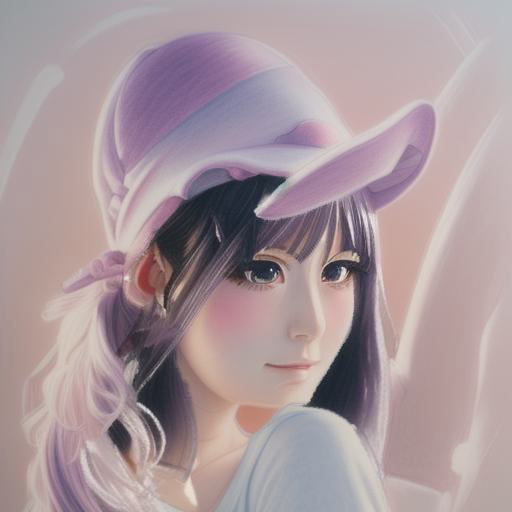}
        \vspace{-1em} %
        \captionsetup{type=figure, skip=0pt}\caption*{-black point}
    \end{minipage}
    \hfill
    \begin{minipage}{0.1625\linewidth}
        \includegraphics[width=\linewidth, height=2.8cm]{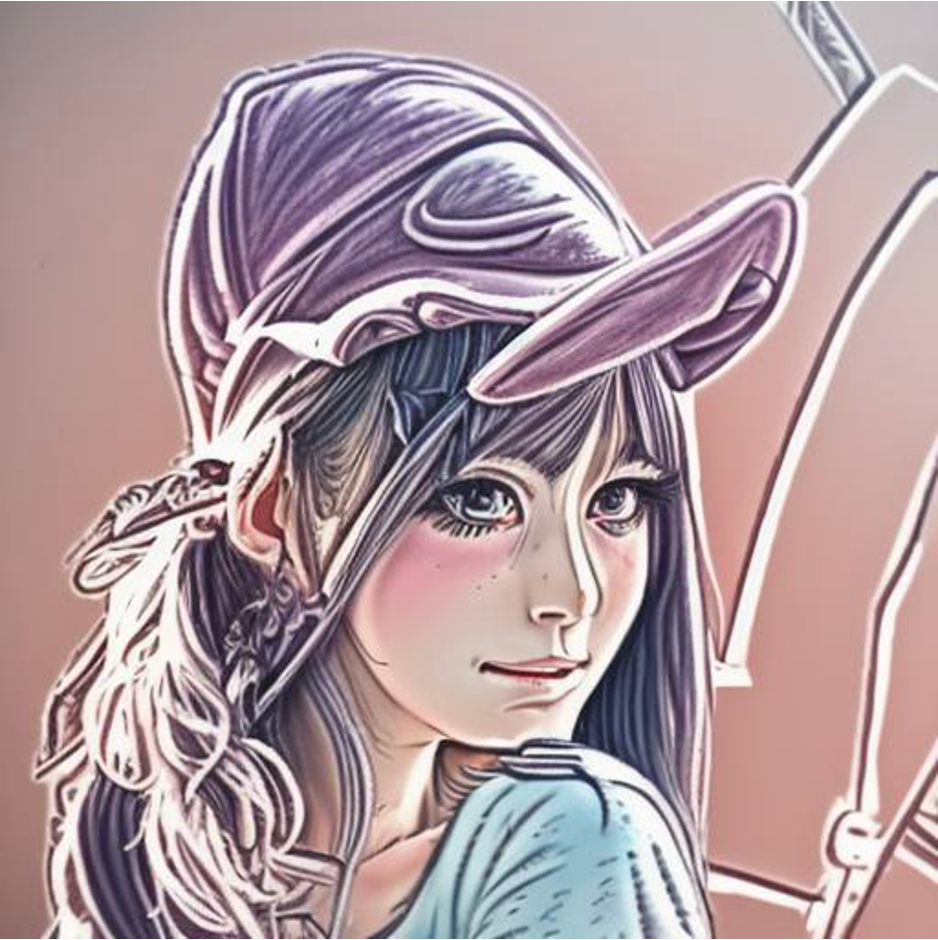}
        \vspace{-1em} %
        \captionsetup{type=figure, skip=0pt}\caption*{+highlights}
    \end{minipage}
    \hfill
    \begin{minipage}{0.1625\linewidth}
        \includegraphics[width=\linewidth, height=2.8cm]{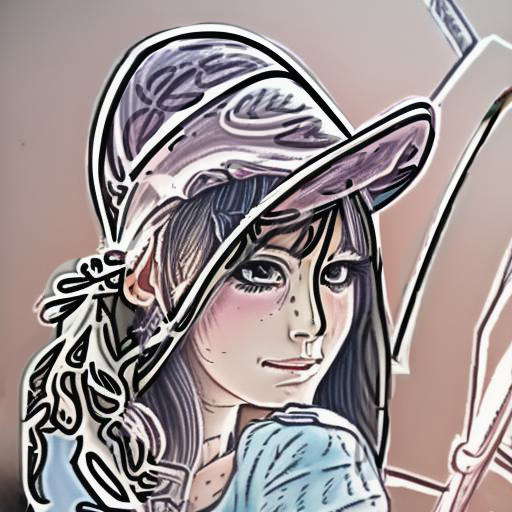}
        \vspace{-1em} %
        \captionsetup{type=figure, skip=0pt}\caption*{+stroke width}
    \end{minipage}
    \hfill
    \begin{minipage}{0.1625\linewidth}
        \includegraphics[width=\linewidth, height=2.8cm]{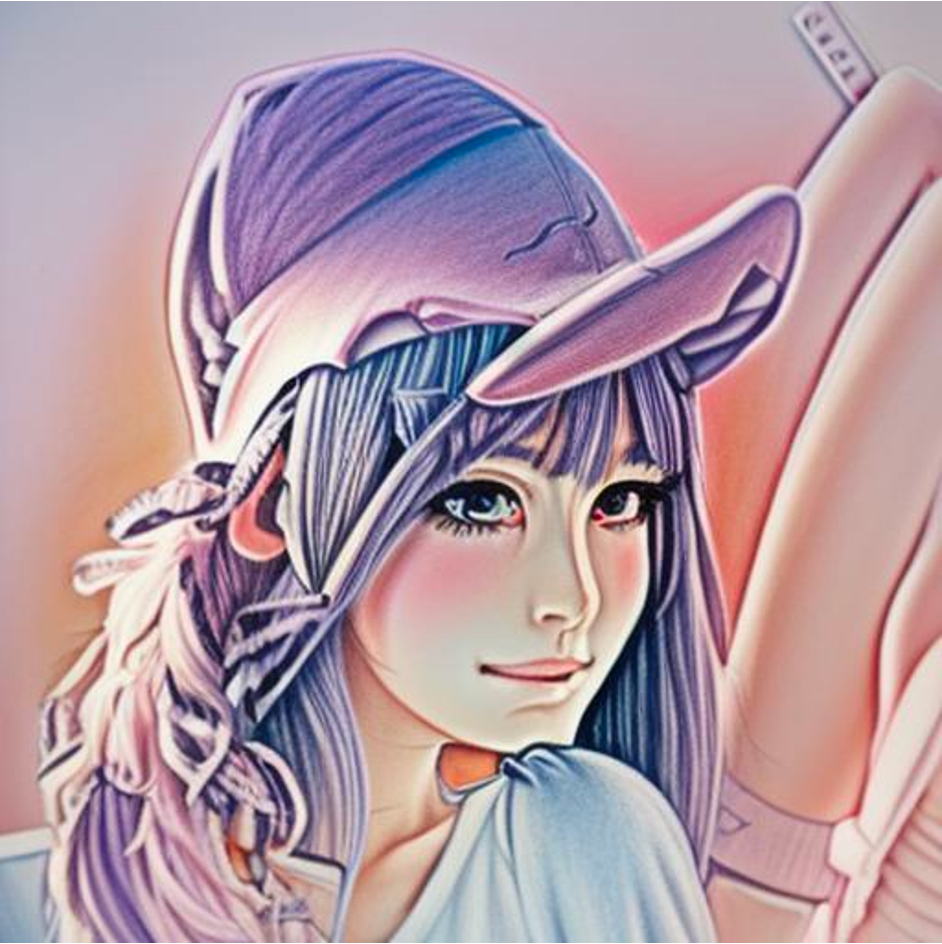}
        \vspace{-1em} %
        \captionsetup{type=figure, skip=0pt}\caption*{+colorfulness}
    \end{minipage}
    \hfill
    \begin{minipage}{0.1625\linewidth}
        \includegraphics[width=\linewidth, height=2.8cm]{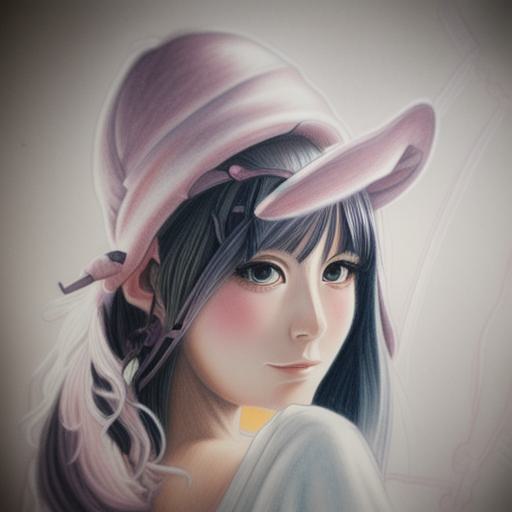}
        \vspace{-1em} %
        \captionsetup{type=figure, skip=0pt}\caption*{+realism}
    \end{minipage}
    \begin{minipage}{0.33\linewidth} %
        \includegraphics[trim={0px 0px 4pc 4px}, clip, width=0.492\linewidth, height=2.8cm]{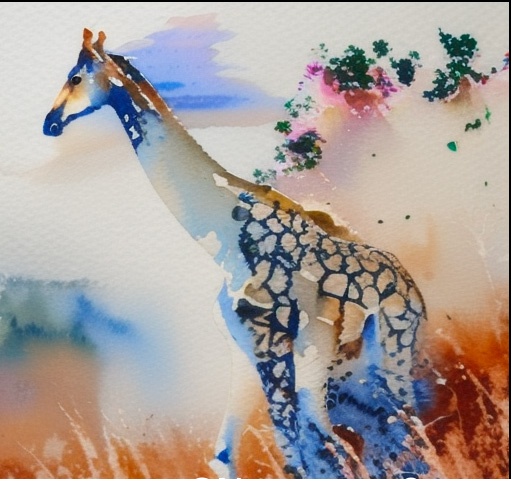}
        \includegraphics[trim={0px 0px 4pc 4px}, clip,width=0.492\linewidth, height=2.8cm]{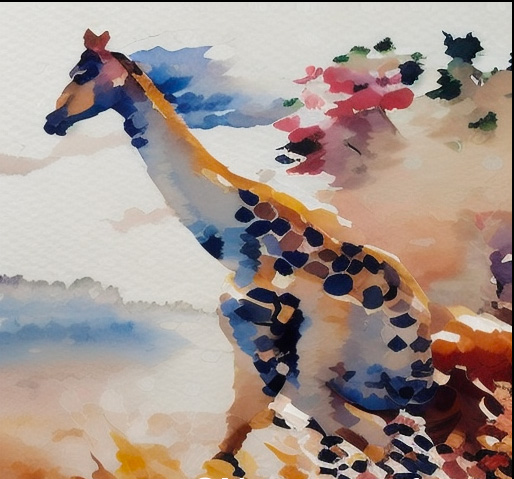}
        \vspace{-1em} %
        \captionsetup{type=figure, skip=0pt}\caption*{Increase watercolor splattering}
    \end{minipage}
    \hfill
    \begin{minipage}{0.33\linewidth}
        \includegraphics[width=0.4925\linewidth, height=2.8cm]{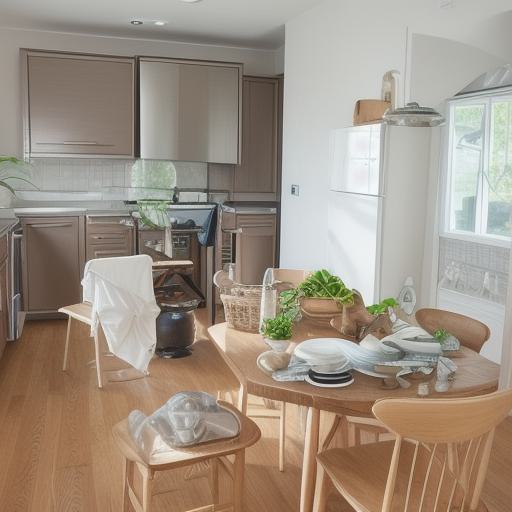}
        \includegraphics[width=0.4925\linewidth, height=2.8cm]{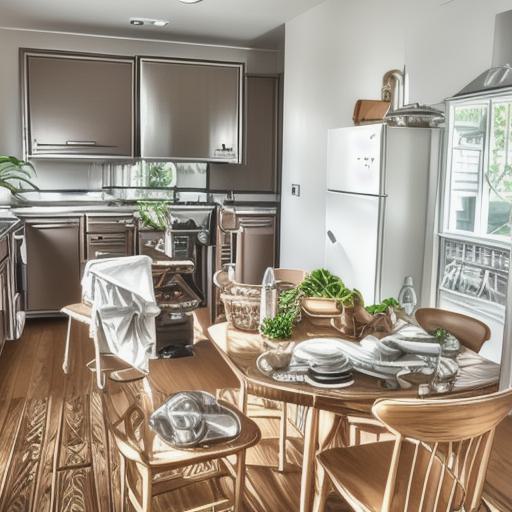}
        \vspace{-1em} %
        \captionsetup{type=figure, skip=0pt}\caption*{Increase local contrast}
    \end{minipage}
    \hfill
    \begin{minipage}{0.33\linewidth}
        \includegraphics[width=0.4925\linewidth, height=2.8cm]{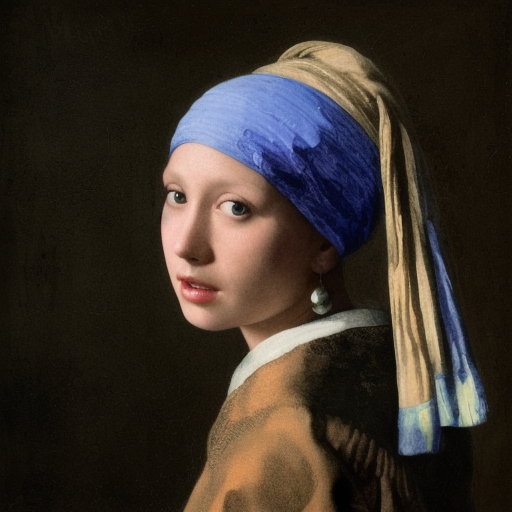}
        \includegraphics[width=0.4925\linewidth, height=2.8cm]{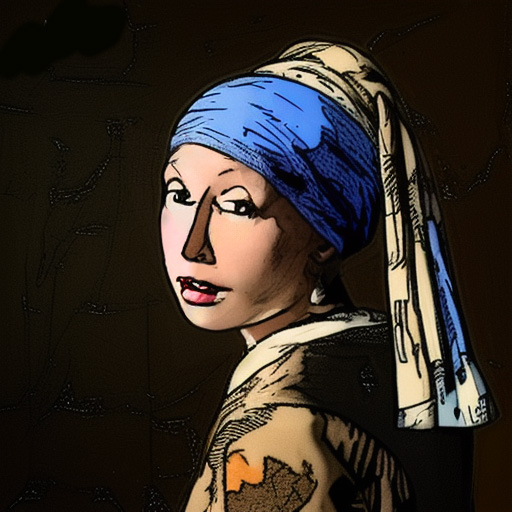}
        \vspace{-1em} %
        \captionsetup{type=figure, skip=0pt}\caption*{Editing of a real image}
    \end{minipage}
\end{minipage}

\captionof{figure}{Our method enables editing of individual stylistic attributes in LDM \cite{rombach2022high} generated images using a stylistic control network. Editing parameters can be combined and adjusted on a continuous spectrum. Using image inversion, real images can be edited as well.  \vspace{2em}}
\label{fig:teaser}
}

\maketitle

\begin{abstract}
  Text-to-image diffusion models have revolutionized image synthesis and editing, but precise control over stylistic attributes remains a challenge, often causing unintended content modifications. We propose an approach for fine-grained parametric control of stylistic attributes in latent diffusion models by learning disentangled editing directions from synthetic datasets. We use guidance composition to close the domain gap between stylistically finetuned and foundation models, preserving the original image semantics while applying stylistic adjustments. To ensure consistent edits, we introduce a training regularization loss and enhance DDIM inversion with optimized null-conditional embeddings for real image editing.  We validate our approach by learning from stylistically filtered synthetic datasets varying a range of stylistic attributes, including outlines, local contrast, watercolorization effects, and geometric patterns. Our evaluations demonstrate that compared to current text-based editing techniques, our method offers well-integrated, more precise and continuously adjustable stylistic modifications.

  \end{abstract}

\FloatBarrier

\input{introduction}

\input{relatedwork}

\input{method}

\input{experiments}

\input{results}
\input{conclusions}

{\small
\bibliographystyle{eg-alpha-doi}  
\bibliography{egbib}
}

\clearpage

\renewcommand*{\thesection}{\Alph{section}}
\setcounter{section}{0}
\renewcommand*{\theHsection}{chX.\the\value{section}}
\noindent {\huge \textbf{Appendix}} \par
\input{supplemental_content}

\end{document}

%% file: acronym.tex
\begin{acronym}
\acro{CG}{Computer Graphics}
\acro{CNN}{Convolutional Neural Network}
\acro{CPU}{Central Processing Unit}
\acro{CV}{Computer Vision}
\acro{DL}{Deep Learning}
\acro{DNN}{Deep Neural Network}
\acro{EPE}{Endpoint Error}
\acro{FPS}{Frames per second}
\acro{GPU}{Graphics Processing Unit}
\acro{GAN}{Generative Adversarial Network}
\acro{K}{Kilo, thousand}
\acro{M}{Million}
\acro{MB}{Megabyte}
\acro{ReLU}{Rectified Linear Unit}
\acro{SSIM}{Structural Similarity Index Measure}
\acro{PSNR}{Peak Signal-to-Noise Ratio}
\acro{FID}{Fréchet Inception Distance}
\acro{LPIPS}{Learned Perceptual Image Patch Similarity}
\acro{MPJPE}{mean per-joint point error}
\acro{MPJPE-PA}{Mean Per Joint Position Error after Procrustes Analysis}
\acro{PVET}{Per-Vertex-Error in T-Pose}
\acro{KID}{Kernel Inception Distance} 
\acro{IS}{Inception Score}
\acro{LPIPS}{Learned Perceptual Image Patch Similarity}
\acro{LDM}{Latent Diffusion Model}
\acro{DM}{Diffusion Model}
\acro{SD}{Stable Diffusion}
\acro{XDoG}{eXtended difference-of-Gaussians}
\acro{NST}{Neural Style Transfer}
\end{acronym}

%% file: introduction.tex
\section{Introduction}
\label{sec:intro}

\begin{figure}[t]
    \centering
    \begin{subfigure}{0.325\linewidth}
        \includegraphics[width=\linewidth]{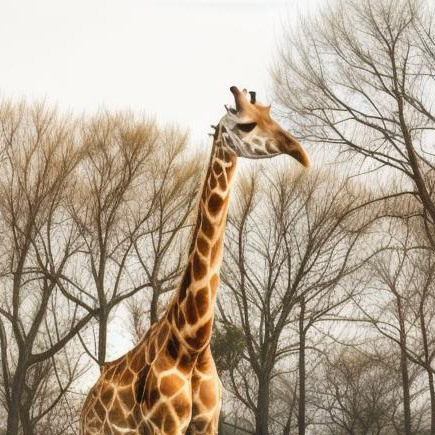}
        \caption{Input}
        \label{fig:contours_xdog_comparison:a}%
    \end{subfigure}
    \hfill
    \begin{subfigure}{0.325\linewidth}
        \includegraphics[width=\linewidth]{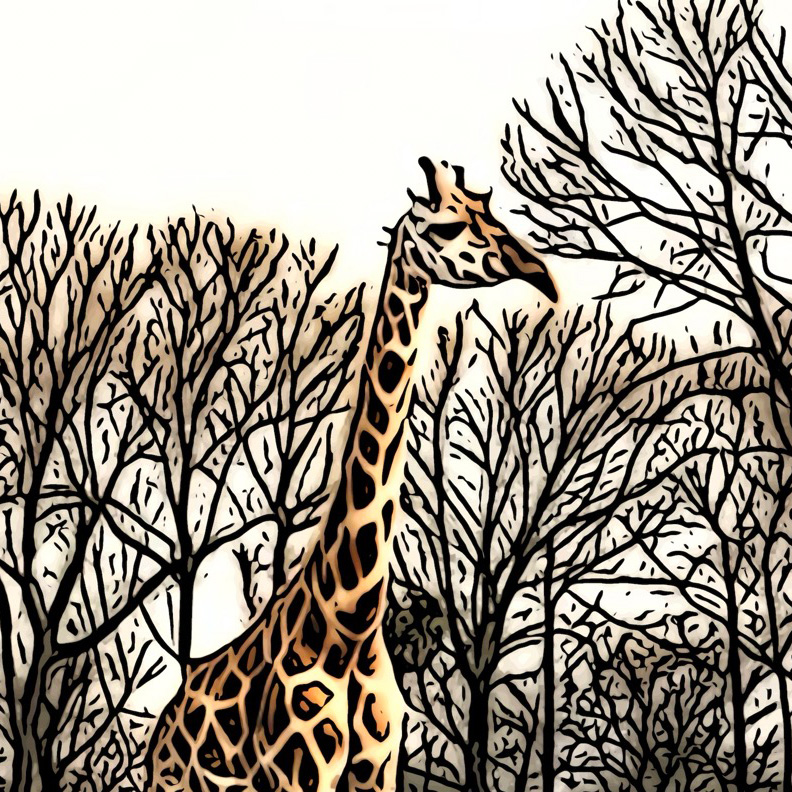}
        \caption{Filtered}
        \label{fig:contours_xdog_comparison:b}%
    \end{subfigure}
    \hfill
    \begin{subfigure}{0.325\linewidth}
        \includegraphics[width=\linewidth]{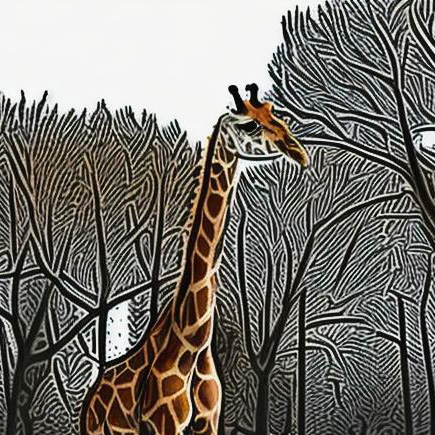}
        \caption{Ours}
        \label{fig:contours_xdog_comparison:c}%
    \end{subfigure}
    \caption{Comparison of xDoG \cite{winnemoller2012xdog} filtered LDM-generated input image  with our xDoG-finetuned method.}
    \label{fig:contours_xdog_comparison}
\end{figure}

Generative diffusion models, especially latent diffusion models (LDMs) such as \ac{SD} \cite{rombach2022high}, have revolutionized image synthesis, producing remarkably detailed outputs from text prompts \cite{rombach2022high,betker2023improving}. Text-to-image models learn a vast range of styles from large-scale paired data, surpassing previous methods in fidelity and diversity. Recent methods like ControlNet \cite{zhangAddingConditionalControl2023} and T2I-Adapter \cite{mouT2IAdapterLearningAdapters2023} improve layout control, while text-based editing \cite{hertz2022prompt,brooksInstructPix2PixLearningFollow2023,zhang2024magicbrush} can replace objects or alter global style. 
However, such methods often struggle at fine-grained editing of stylistic elements like color, texture, and stroke patterns due to prompt ambiguities, unintended content modifications and inability to control edits on a continuous scale. Thus, a Photoshop-like editing workflow, allowing artist to adjust parameter sliders for specific aspects of the style in a controlled manner during image generation is currently not possible.

Earlier filter-based non-photorealistic rendering (NPR) methods\cite{Kyprianidis:2013:SAT}, allow fine grained parameter control (e.g., line width in \cref{fig:contours_xdog_comparison:b}), but they operate on low level features and often fail to capture scene semantics, a shortcoming that has repeatedly been mentioned as motivation for methods such as \ac{NST}~\cite{Gatys2016ImageST,Jing_2020_NST} and content-aware tone pipelines~\cite{gharbi2017deep}.  Moreover, as filters are applied post-synthesis they cannot benefit from shared scene understanding and user intent, such as the text- or geometry-guidance during reverse diffusion. For instance, increasing edges using a NPR-based edge filter thickens all edges equally (\cref{fig:contours_xdog_comparison:b}), whereas our LDM-integrated approach (\cref{fig:contours_xdog_comparison:c}) selectively places stylistic emphasis on object contours and important features (e.g., tree stems).

Our approach introduces a method to apply parametric control over specific stylistic attributes, where attribute strength and location can be precisely controlled in SD generation. 
Such adjustments are difficult to define through text alone and are often entangled in style image references, making traditional style transfer methods unsuitable, yet are easily generated as small synthetic datasets of desired adjustments. For example, varying outline strength and sensitivity in an edge filter \cite{winnemoller2012xdog}, we can fine-tune an adapter for outline control in SD generation  (\cref{fig:contours_xdog_comparison:c}). In \cref{fig:teaser} we demonstrate control over attributes such as brush stroke width, local contrast, highlighting, blackpoint, effect-specific controls such as watercolor paint splattering, as well as real image editing.

However, directly finetuning adapters on desired stylistic attributes often results in domain shifts away from the original \ac{SD} distribution and can degrade output quality and diversity.
To mitigate this, we re-align the output latents to the \ac{SD} domain using guidance composition and introduce a training regularization that discourages undesired semantic changes. 
For real-image editing, we adapt null-text inversion \cite{mokadyNULLTextInversionEditing}, optimizing null-parameter embeddings for stable reconstructions. Our experiments demonstrate that this integrated approach can precisely adjust stylistic attributes without undesired changes to the image semantics or global style. Compared to previous approaches, our edits can furthermore be adjusted on a continuous scale, allowing slider-based fine-grained style control. We plan on releasing the code and pretrained models.

%% file: relatedwork.tex
\section{Related Work}
\label{sec:relatedwork}

\subsection{Traditional Image Filtering Techniques}
In contrast to deep learning, traditional image-based artistic rendering (IB-AR) methods \cite{Kyprianidis:2013:SAT} offer granular control through a series of engineered filters designed for specific artistic styles. Although highly controllable, these methods are constrained to specific, handcrafted styles like cartoon, oilpainting or watercolor effects \cite{winnemoller2006real, semmo2016image, bousseau2006interactive} or stroke-based \cite{liu2021paint,zou2021stylized} or learnable filter-based \cite{lotzsch2022wise, reimann2024artistic} de- and re-composition of images. Further, their post-hoc filtering is not informed of high level semantics and user intent that the text-to-image generative methods are given, such as prompt and spatial guidance, reducing their ability for content- and style aware adjustments. We make use of IB-AR for synthetic data generation of parameter-controllable stylistic attributes, where our fine-tuned controls do not seek to exactly replicate the effects of filter application but to provide an editing direction to the LDM along the range of the filter-parameter.

\subsection{Diffusion-based Control Techniques}
Methods such as ControlNet \cite{zhangAddingConditionalControl2023} and T2I-Adapter \cite{mouT2IAdapterLearningAdapters2023}, have enabled spatial control of \acp{DM} \cite{sohl2015deep, dhariwal2021diffusion} by integrating guidance maps. Many methods explore prompt-based semantic editing of images, like the training-free prompt-to-prompt \cite{hertz2022prompt} and null-text inversion \cite{mokadyNULLTextInversionEditing}, which manipulate the attention and guidance vectors of \acp{LDM} during inference, as well as the model-training based InstructPix2Pix \cite{brooksInstructPix2PixLearningFollow2023}, MagicBrush \cite{zhang2024magicbrush} and, specifically for styles, StyleBooth \cite{han2024stylebooth}, which learn instruction-based editing.
In our work, we make use of ControlNet for spatial guidance and demonstrate that out proposed approach is better suited for disentangled editing of specific stylistic elements like color, texture, and stroke patterns, and is significantly better for continuous scale editing than these instruction-based editing methods.

%% file: method.tex
\section{Method}
\label{sec:method}

\begin{figure*}
    \centering
    \includegraphics[width=0.85\linewidth,trim={1cm 1.2cm 1cm 0.5cm},clip]{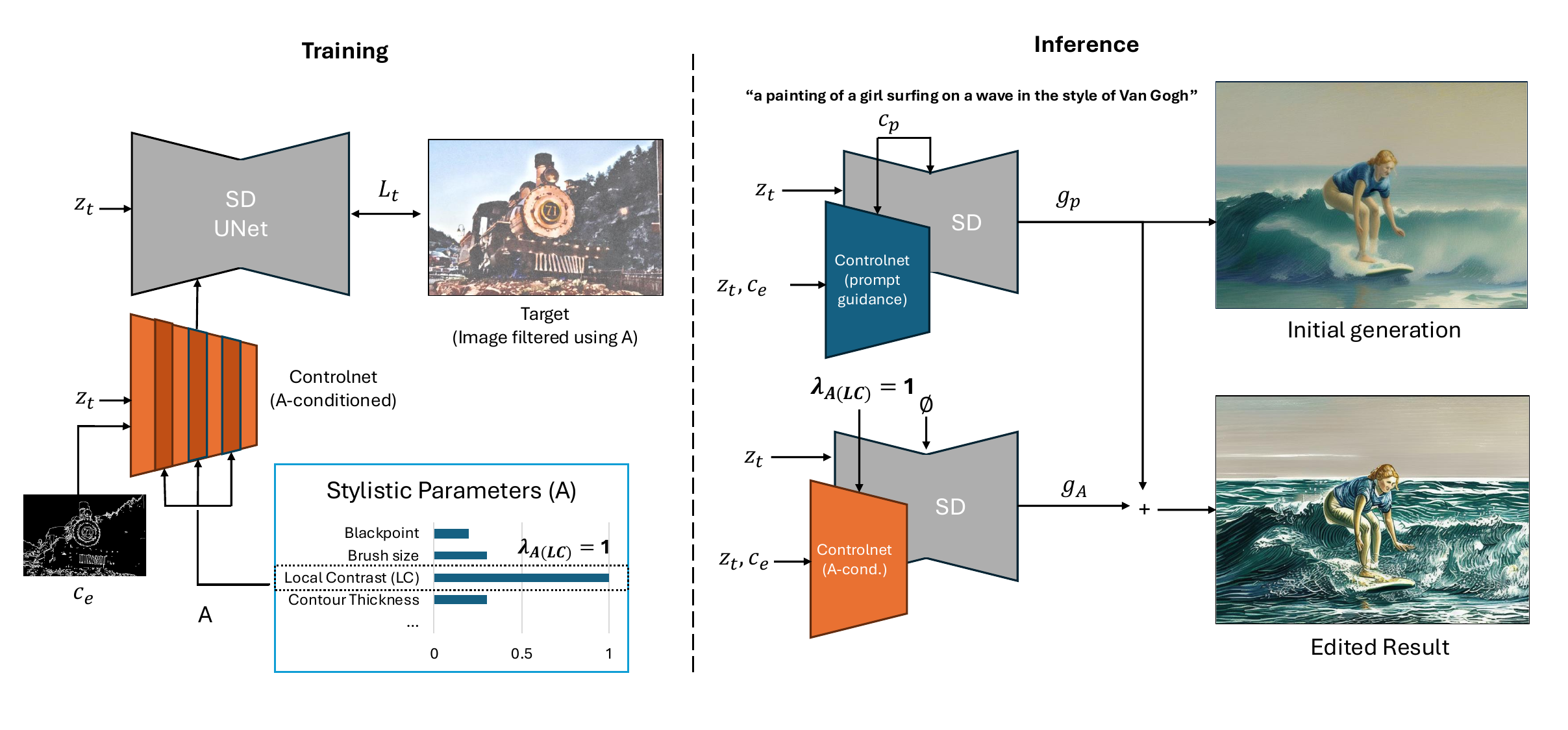}
    \vspace{1em}
    \caption{Overview of our approach. During training, attention layers (dark orange) of an edge-conditioned Controlnet are finetuned to reproduce the images of a synthetic dataset with varied effect settings $\lambda_A$. During inference, the finetuned network acts as a stylistic control to a pretrained Controlnet and LDM (\ac{SD}), modifying the attribute $A$ in the image using guidance $g_A$ while adhering to content and style. }
    \label{fig:overview_graphic}
\end{figure*}

Let $I$ denote an image generated by denoising a latent $z_T$, which is either sampled from random noise or obtained by inversion of a real image. Our goal is to edit $I$ by precisely adjusting specific stylistic attributes $A={A_1, \dots, A_n}$ globally or locally, controlled by continuous parameter strengths $\lambda_{A_1} \dots \lambda_{A_n}$, resulting in an edited image $I^*$. These parameter-based adjustments of global and local image properties must be learned explicitly from a dataset  encoding the desired stylistic attributes with varied strengths $\lambda_A$, as textual prompts alone lack the precision required for such edits.
Naively fine-tuning the diffusion model directly on such datasets however results in a domain shift characterized by reduced diversity and compromised generative quality due to the comparatively small and less content-diverse target datasets. We formalize this as a \emph{covariate shift}, where the original Stable Diffusion model  defines a source domain distribution $p_{\mathcal{SD}}(x \mid c)$ conditioned on text prompts $c$, and the dataset of stylistically edited images defines a narrower target domain distribution $p_{\mathcal{T}}(x \mid c, \lambda_A)$.
To address this covariate shift, we finetune a LDM $\varepsilon_A$ (we follow standard LDM~\cite{rombach2022high} notation, see suppl. for background) designed to steer image generation toward the target stylistic attributes, and extract the attribute representations using a guidance formulation at inference. 
To maintain alignment with the source domain distribution, the isolated stylistic guidance from $\varepsilon_A$ is composed with the original diffusion model output $\varepsilon_\theta$ at each denoising step of the reverse diffusion process. 

In the following, we detail our guidance-based composition strategy (\cref{subsec:method:effectguidance}), approach to maintain spatial stability (\cref{subsec:method:spatialstability}), and network training approach (\cref{{subsec:method:network}}). Furthermore, we extend this technique to real image editing through null-embedding inversion outlined in \cref{subsec:method:real_image_editing}.

\subsection{Effect Guidance}
\label{subsec:method:effectguidance}
After fine-tuning on an effect-specific dataset, we isolate the learned stylistic control $A$ in  $\varepsilon_A$ from layout, semantic and other stylistic appearance features inherent in the training data. To this end, we define a editing vector $g_A$ towards our desired edit $A$ by
 encoding $g_A(\lambda_A=k)$ as the difference between noisy predictions at the desired stylistic strength $k$ and the neutral baseline ($\lambda_A=0$):

\begin{equation}
g_A = \varepsilon_{A}(z_t,t,\emptyset,\lambda_A=k) - \varepsilon_{A}(z_t,t,\emptyset,\lambda_A=0)
\end{equation}

Note that we do not supply prompt embeddings ($\emptyset$) to both networks.
Since guidance vectors are compositional \cite{liu2022compositional}, we can then introduce our fine-tuned guidance vector to the text-conditioned base SD ($\varepsilon_{\theta}$) classifier-free guidance \cite{hojonathanClassifierFreeDiffusionGuidance2022}  as:
\begin{equation}
\varepsilon_{\theta}(z_t,t,g_p,g_A) = \varepsilon_{\theta}(z_t,t,\emptyset) + w_1 g_{p} + w_2 g_A
\end{equation}

where $w_1,w_2$ are classifier free guidance scales. 

\subsection{Spatial Stability}
\label{subsec:method:spatialstability}

\begin{figure}
    \centering
    \includegraphics[width=1\linewidth]{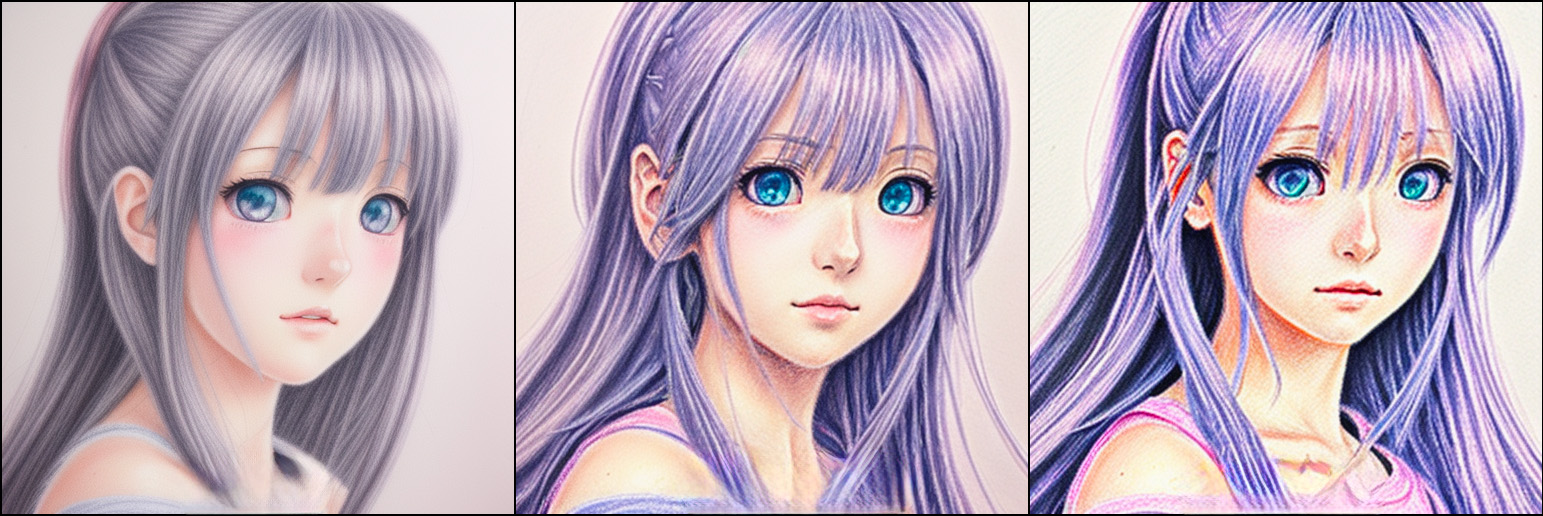}
    \caption{Using our stylization guidance ($g_{A}$) without Controlnet guidance. Here we increase the linewidth parameter. While results vary in line strength, they are not spatially stable. }
    \label{fig:stylization_no_controlnet}
\vspace{-1em}
\end{figure}

While applying the above formulation to an \(\varepsilon_\theta\) conditioned solely on text can yield visually compelling results (\cref{fig:stylization_no_controlnet}), the outputs often lack spatial stability. Minor variations early in the diffusion process can lead to significantly different image layouts. To address this, we fix the coarse spatial layout by conditioning both \(\varepsilon_\theta\) and \(\varepsilon_A\) on spatial maps, utilizing a ControlNet $C$ \cite{zhangAddingConditionalControl2023}. We use Canny edge image $c_e$ conditioned ControlNets as spatial anchors due to their generality and ease of creation. Specifically, \(\varepsilon_\theta\) is steered using a pretrained Controlnet, receiving $C(c_p,c_e)$, while \(\varepsilon_A\) is steered using our A-finetuned Controlnet $C(A,c_e)$ (see \Cref{fig:overview_graphic}). The predicted noise using classifier guidance is thus actually 
\begin{equation}
\varepsilon_{\theta}(z_t,t, c_e, g_p, g_A) = \varepsilon_{\theta}(z_t,t,\emptyset,C(c_p,c_e)) + w_1 g_{p,e} + w_2 g_{A,e}
\label{eq:full_guidance_w_cnet}
\end{equation}
where $g_{p,e}$ and $g_{A,e}$ represent the ControlNet-conditioned prompt and attribute guidance vectors.
Spatial conditioning significantly enhances consistency and reduces unintended correlations between layout and stylistic effects. However, regions lacking clear spatial features (e.g., without edges) may still experience unintended semantic alterations. To mitigate such content drift, we activate stylistic guidance $g_{A,e}$ after timestep $\text{act}_t$, restricting semantic freedom of $\varepsilon_{A}$. Formally, we set guidance activation as $w_2(t) = 0$ for $t_{\text{norm}} < \text{act}_t$, with $t_{\text{norm}}$ normalized between 0 (start) and 1 (end) of denoising. We empirically set $\text{act}_t=0.1$ to balance effect strength and stability, further complemented by a training regularization term detailed in~\cref{eq:training_reg}.

\subsection{Network}
\label{subsec:method:network}

\paragraph*{Architecture.} We use a ControlNet architecture $C_A$ and insert a layer with an extra attention transformer-block after the text-conditional cross attention into the overall downsampling blocks. The extra attention layer receives an embedding of the input parameters $\lambda_A$. During training, only the extra attention layers and the $\lambda_A$-embedder are finetuned, preserving the original weights of the LDM.  Ablation studies (\cref{subsec:ablationstudies}) show that this variant keeps the results more stable compared to finetuning the cross attention blocks or encoding parameters into tokens.

\paragraph*{Training Regularization.}
During inference, changes introduced into the image often get more drastic with increased parameter values $\lambda_A$, which also includes unwanted semantic changes. These are encoded in the guidance term $g_A$  between the zero-conditional and the parameter term. To combat this, we introduce a regularization term. Intuitively, the regularization aims at penalizing the difference of images produced with $\lambda_A=0$ and $\lambda_A>0$  at spatial locations that do not correspond to changes in the input training images. Let $z_{t-1,\lambda=k}$ be a latent sample obtained from noising the image $I_{A,\lambda=k}$ of effect strength $k$ after one step of DDIM~\cite{song2020score}. Further, let $z_{t,0}$ be the latent where the effect was not applied. 
The regularization loss is then defined as:
\begin{equation}
 \mathcal{L}_{reg} = \left\|\frac{|z_{t-1,\lambda=k} - z_{t-1,\lambda=0}|}{1 + |z_{0,\lambda=k} - 
 z_{0,\lambda=0}|}\right\|^2_2.
 \label{eq:training_reg}
\end{equation}

\paragraph*{Training Loss.}
We finetune the ControlNet attention blocks using the well-known diffusion training strategy \cite{song2020denoising, dhariwal2021diffusion, rombach2022high}. We extend the training loss of ControlNet~\cite{zhangAddingConditionalControl2023} to incorporate our conditioning on parameter $A$, fixed guidance maps $c_e$ and the training regularization, which is controlled with strength $\beta$. The training loss for a timestep $t$ is then defined as:
\begin{equation}
L_t = \lVert \varepsilon-\varepsilon_{A}(z_t, t, c_{\text{p}},c_{\text{e}},\lambda_A) \rVert_{2}^{2} + \beta\mathcal{L}_{reg}
\end{equation}

\subsection{Real image editing}
\label{subsec:method:real_image_editing}
Let $I$ be a real image. To achieve image editing of $I$, we have to invert $z_0=D(I)$ into noise $z_T$. DDIM inversion \cite{dhariwal2021diffusion} reverses the diffusion process, transforming an encoded real image  $z_0 $ back into a noise vector $z_T$ through incremental steps of reversed DDIM. However, DDIM inversion alone is not sufficient to capture good editing directions for prompt-based editing \cite{mokadyNULLTextInversionEditing}, as inversed latents $z_T$ can diverge from their original trajectory due to the guidance scale amplifying reconstruction errors, resulting in bad editing results. To remedy this, Mokady et al. \cite{mokadyNULLTextInversionEditing} propose to also optimize the null-text embedding $\emptyset_t$ during inversion.
The optimized null text embedding hereby plays the role of a pivot towards a more editable direction in CfG latent space.
First, the DDIM inversion is performed with CfG guidance scale $w = 1$ and outputs a sequence of initial latent codes $\tilde{z}_T^*, \ldots, \tilde{z}_0^*$. 
These are then used as targets for the optimization of the null text embedding $\emptyset_t$, minimizing the error between $z^*_t$ and $z_{t-1}(z_t, \emptyset_t, c_p)$, which is the latent code obtained from a step of DDIM sampling with guidance $w=7.5$. This procedure is repeated in each timestep, initialising $z_{t-1}$ and $\emptyset_{t-1}$ with the previous step results, and yielding $\{\emptyset_t\}_{t=1}^T$ optimized embeddings. Please refer to Mokady et al.~\cite{mokadyNULLTextInversionEditing} for details on the algorithm.

We adapt this approach, but instead of optimizing the null-\emph{text} embedding, we optimize the null-parameter ($\lambda_A=0$) embedding $\emptyset^A_t$ used in $g_{A}$. Further, compared to the original formulation, we invert \cref{eq:full_guidance_w_cnet}, including the applications of the two controlnets. 
Thus, for every timestep $t$ we optimize for N iterations
\begin{equation}
\min_{\emptyset^A_t} \left\| \tilde{z}_{t-1}^* - z_{t-1}(\tilde{z}_t, \emptyset, \emptyset^A_t, c_e, c_p, \lambda_A) \right\|_2^2. 
\end{equation}
During image editing inference, the stored noise latent $\tilde{z}_T$ and optimized null-parameter conditional embeddings $\{\emptyset^A_t\}_{t=1}^T$ can then be injected to edit the parameters. 
Joined optimization of both text and null conditions leads to degraded results, as we show in an ablation experiment (\cref{subsec:ablationstudies}).

%% file: experiments.tex
\section{Experiments}
\label{sec:experiments}

\begin{figure}[t]
    \centering
    \begin{subfigure}[b]{0.24\linewidth}
        \centering
        \includegraphics[width=\linewidth]{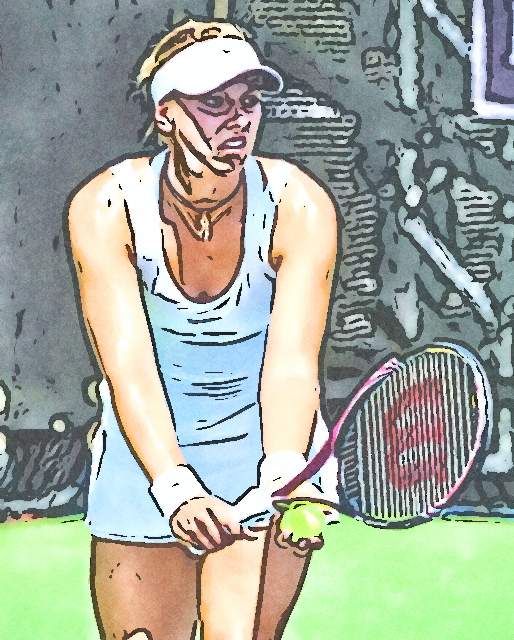}
        \caption{\footnotesize line width=1}
        \label{fig:contourwidth_100}
    \end{subfigure}
    \hfill
    \begin{subfigure}[b]{0.24\linewidth}
        \centering
        \includegraphics[width=\linewidth]{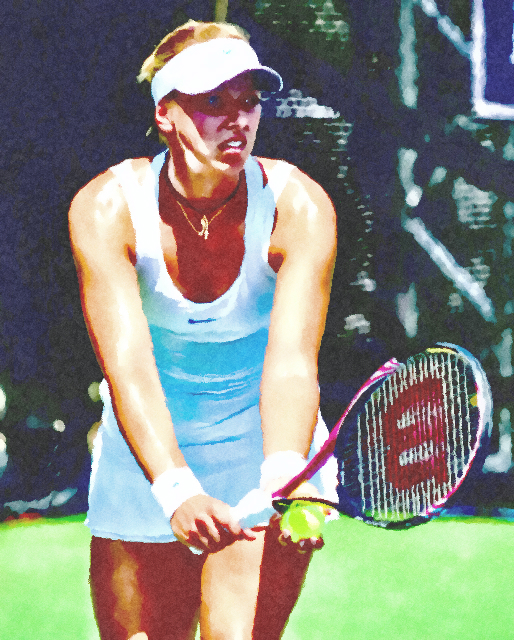}
        \caption{\footnotesize black point=1}
        \label{fig:blackpoint_100}
    \end{subfigure}
    \hfill
    \begin{subfigure}[b]{0.24\linewidth}
        \centering
        \includegraphics[width=\linewidth]{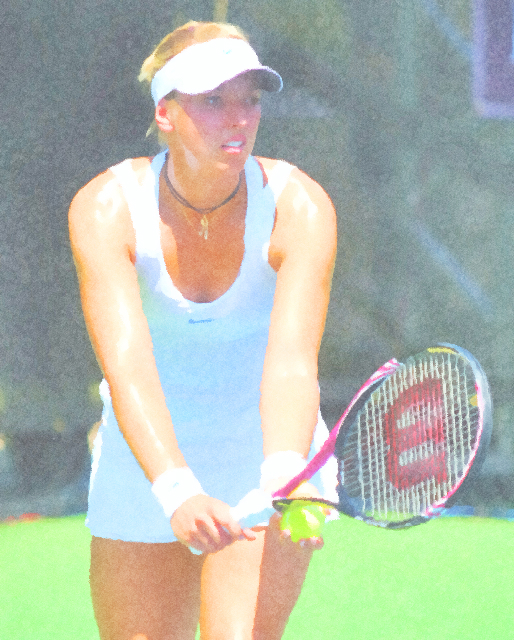}
        \caption{\footnotesize depth pow.=0}
        \label{fig:depthPower_0}
    \end{subfigure}
    \hfill
    \begin{subfigure}[b]{0.24\linewidth}
        \centering
        \includegraphics[width=\linewidth]{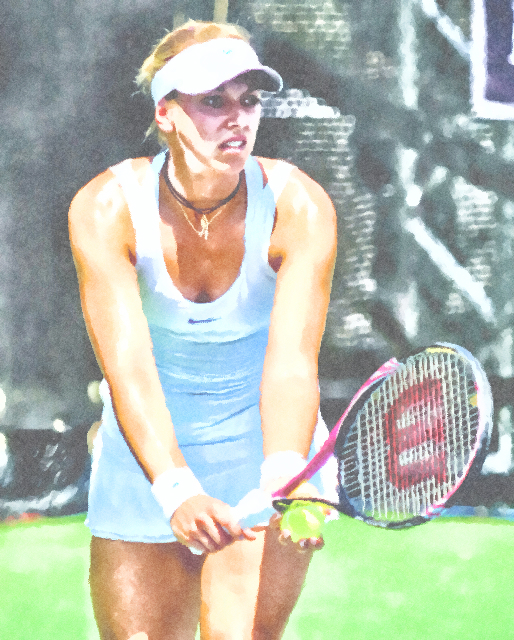}
        \caption{\footnotesize depth pow.=1}
        \label{fig:depthPower_100}
    \end{subfigure}
    \caption{Examples of training images stylized with watercolor filter for different parameter settings.}
    \label{fig:training_examples}
\end{figure}

\begin{figure}[t]
    \centering
    \begin{subfigure}[b]{0.45\linewidth}
        \centering
        \includegraphics[width=\linewidth]{graphics/training_examples/PTT_dog.jpg}
        \caption{\footnotesize PTT small strokes}
        \label{fig:PTT_0}
    \end{subfigure}
    \begin{subfigure}[b]{0.45\linewidth}
        \centering
        \includegraphics[width=\linewidth]{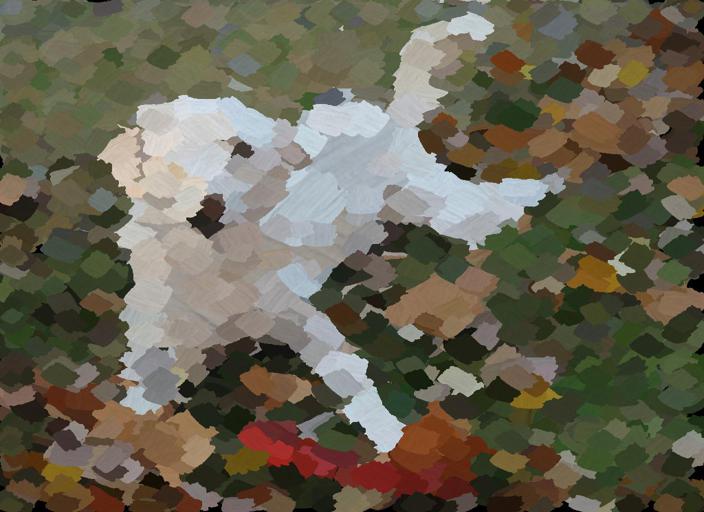}
        \caption{\footnotesize PTT large strokes}
        \label{fig:PTT_100}
    \end{subfigure}
    \caption{Examples from PaintTransformer \cite{liu2021paint} }
    \label{fig:ptt_example}
\end{figure}

\subsection{Training Dataset}  
\paragraph*{Synthetic.} We generate synthetic training datasets by creating multiple stylized renditions of content images with a stylization filter $F$, varying the strengths of its parameters $\lambda_{A(k)} \in (0, 1)$ for each image. Our variations concern only specific attributes of the effect, not the overall effect intensity, i.e., at $\lambda_{A(k)} = 0$, the stylization effect (e.g., watercolorization) remains visible. We generate the following datasets, using  MSCOCO train2014~\cite{MSCOCO} as content images:

\begin{itemize}
    \item \textbf{Watercolor.} We use a watercolorization filter~\cite{bousseau2006interactive}, adjusting seven parameters:  ``contourWidth" (\cref{fig:contourwidth_100}) and ``details" for controlling outlines and fine brushstrokes; ``colorfulness" and ``blackpoint" (\cref{fig:blackpoint_100}) for histogram adjustments; watercolor-specific effects such as  “paintSplatter” and ``wobbling"; and ``depthPower" (\cref{fig:depthPower_0,fig:depthPower_100}) for adaptive contrast enhancement~\cite{gharbi2017deep}.
    \item \textbf{Cartoon.} We employ a cartoonization filter~\cite{winnemoller2006real}, which combines color quantization with thick outlines generated by \ac{XDoG}. 
    \item \textbf{Stroke Rendering.} We utilize PaintTransformer~\cite{liu2021paint} using coarse brushstrokes to create a stroke-based aesthetic (\cref{fig:ptt_example}).
\end{itemize}

\paragraph*{Real Artworks.} We train on Artbench \cite{liao2022artbench}, containing real artworks for different styles, and create photographic versions of each using a Controlnet conditioned on the artwork canny edge maps. This allows us to vary the realism scale of stylized images (see \cref{fig:teaser}, right-most image).

\subsection{Ablation Studies}
\label{subsec:ablationstudies}
\begin{figure}[t]
    \centering
    \begin{subfigure}{0.49\linewidth}
        \includegraphics[width=\linewidth]{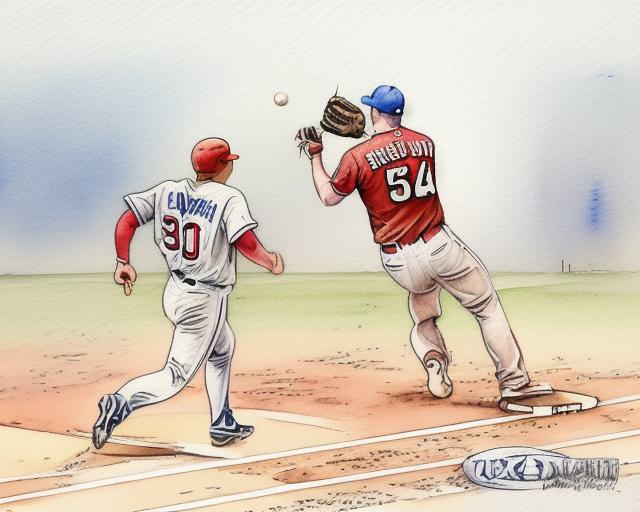}
        \caption{5k steps w/o reg.}
    \end{subfigure}\hfill
    \begin{subfigure}{0.49\linewidth}
        \includegraphics[width=\linewidth]{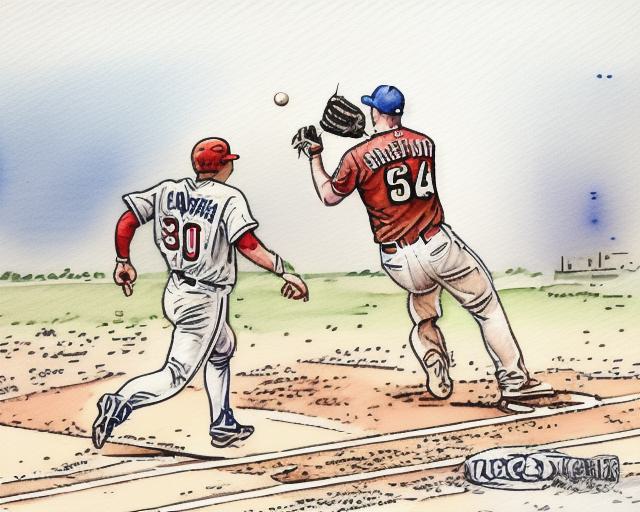}
        \caption{5k steps $\beta=5$}
    \end{subfigure}
    \caption{Comparison of training outputs for $\lambda_{\text{strokewidth}}=1$ with and without regularization. Regularization leads to early convergence.}
    \label{fig:comparison_training_progress}
\end{figure}

\begin{figure}[t]
    \centering
    \begin{subfigure}{0.32\linewidth}
        \includegraphics[width=\linewidth]{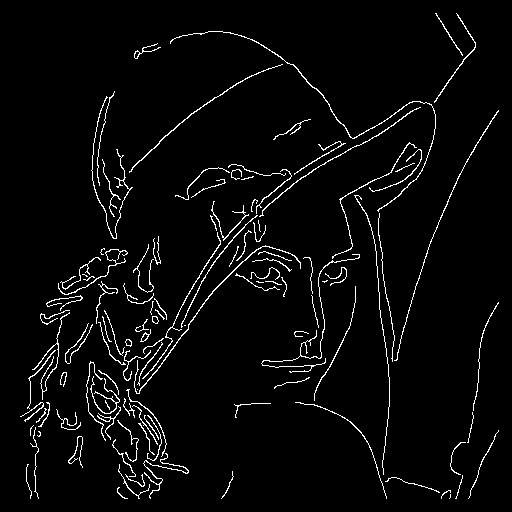}
        \caption{canny edge map}
    \end{subfigure}
    \begin{subfigure}{0.32\linewidth}
        \includegraphics[width=\linewidth]{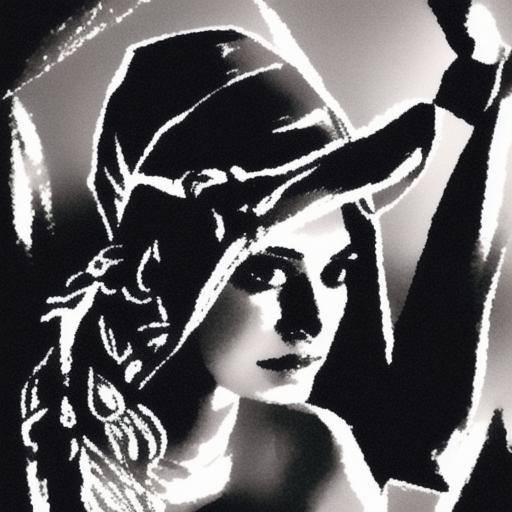}
        \caption{black point=1}
    \end{subfigure}
    \begin{subfigure}{0.32\linewidth}
        \includegraphics[width=\linewidth]{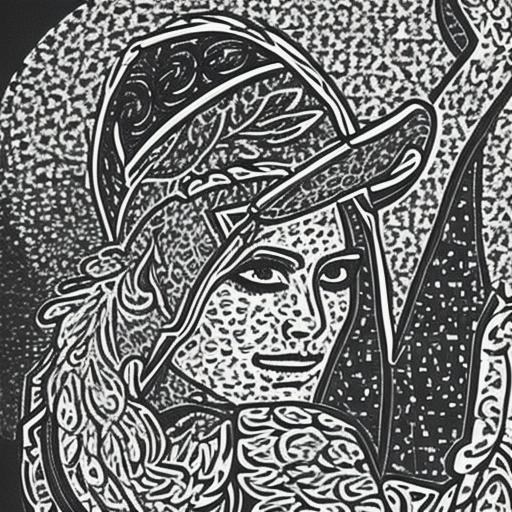}
        \caption{details=1}
    \end{subfigure}
    
    \caption{Outputs of only $\varepsilon_A$. Finetuned layers, conditioned on the edge map (a), learn to discard most colour and scene information. }
    \label{fig:cnet_only_out}
\end{figure}

\paragraph*{Without guidance.}
In \cref{fig:cnet_only_out}, we present outputs from the finetuned Controlnet alone (i.e., $\varepsilon_{A,\lambda_A=0} + w_2g_A$), showing that training extra attention layers conditions model outputs on parameter-specific changes. Our approach is thus able to disentangle and apply individual style attributes to new domains (defined by the \ac{SD} prompts at inference), without reproducing the overall style or content of the training images.

\paragraph*{Testing dataset.}
To evaluate our model, we created a test dataset ($D_T$) with stylized and edited images. We obtained base prompts (using BLIP~\cite{li2022blip} captioning) and canny edge maps on the first 50 images of MSCOCO val2014. Using these, we generate images in photographic rendition and stylized version (Van Gogh and watercolor style) using prompt modifiers. For each base prompt, baseline images $D_{T0}$ were generated using the baseline ControlNet model ($\lambda_A=0$), while edited images $I_A$ were produced using our approach ($\lambda_A=1$). Reference images $I_R$ were generated by applying the style filter to baseline images: $I_R=F_A(I_0 \in D_{T0},\lambda_A=1)$, providing approximate ground truths for comparison.
We conducted experiments using the watercolor training dataset, specifically evaluating parameters: \textit{contourWidth} (local structure), \textit{depthPower} (global/local contrast), and \textit{details} (fine local brushstroke-like effects). Prompt modifiers included “a watercolor painting” for watercolor style, “a painting by Van Gogh” for painterly style, and no modifier for photographic style.

\paragraph*{Parameter Embedding.}

\begin{figure}[ht]
    \centering
    \includegraphics[width=\linewidth]{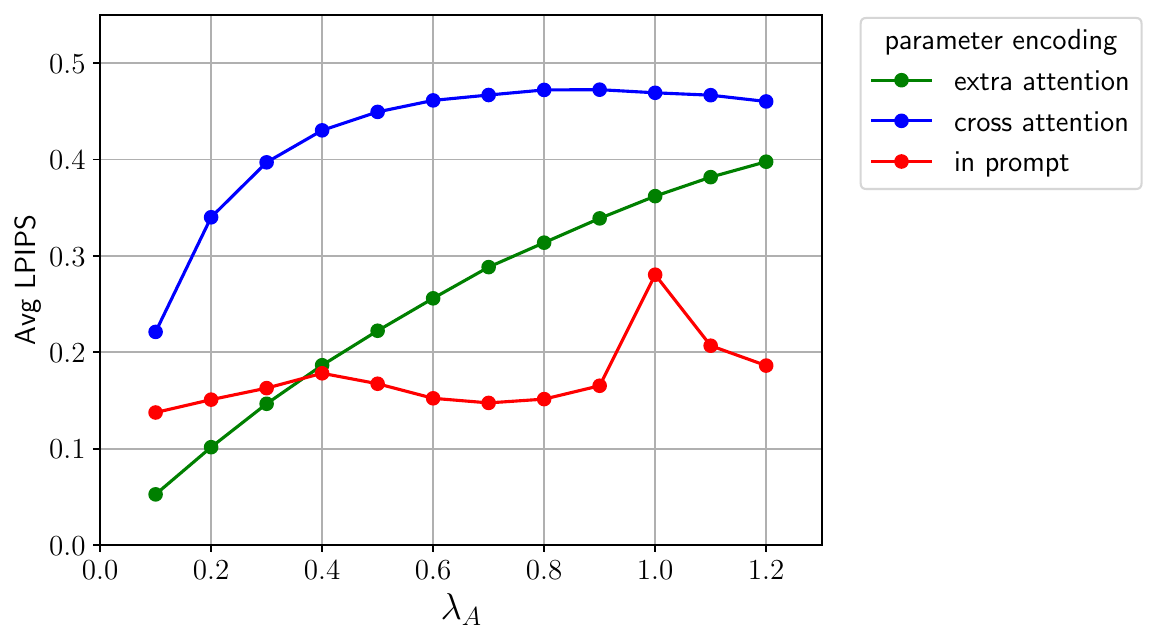}
    \caption{Parameter encoding ablation. We measure LPIPS of outputs to the unedited original as $\lambda_A$ (depth power of the watercolor filter) increases. Using an extra attention layer (our method) provides a stable response across the parameter range.}
    \label{fig:parameter_encoding_variation}
\end{figure}

We ablate the parameter embedding method, exploring alternative parameter incorporation strategies during finetuning and inference. We tested two configurations, both finetuning the original text cross-attention layers instead of our proposed extra attention layer. The first variant also uses a linear embedder, directly mapping input parameters into the cross-attention dimension, omitting explicit text conditioning already provided by $g_{p,e}$ in \cref{eq:full_guidance_w_cnet}. The second variant preserves textual context by appending parameter descriptions to BLIP-generated captions~\cite{li2022blip} (e.g., ``stylized with [parameter] set to [value*100]\%").
In \cref{fig:parameter_encoding_variation} we vary $\lambda_{\text{depth-power}} \in [0,1.2]$ on generating $D_T$. Cross-attention finetuning yields plausible but abrupt transitions at lower parameter values and plateaus thereafter, often causing content changes. Prompt-embedded parameters exhibit limited smoothness and fail to interpolate effectively. Conversely, our extra attention method achieves consistent and smooth control, also outside the training range ($[0.0,1.0]$), demonstrating superior stylistic stability and gradual parameter adjustment.

\paragraph*{Training and Regularization.}
Regularization enhances the difference between $\lambda = 0$ and $\lambda = 1$, acting as an amplifier of local changes. \cref{fig:comparison_training_progress} and supplemental materials compare outputs from 5k steps with and without regularization, demonstrating that regularization achieves the same stylization level at 5k steps as unregularized training at 30k steps. However, more training steps can introduce unintended effects, such as background smoothing at 30k steps. Comparisons with $I_R$ (see supplemental) show that higher regularization increases divergence from algorithmic filtering, with high regularization potentially making stylizations overly strong. We found that setting regularization to $\beta=5$ and using 5k steps strikes a balance between quality stylization and training efficiency.

\begin{figure}
    \centering
        \begin{subfigure}{0.24\linewidth} 
    \includegraphics[width=\linewidth]{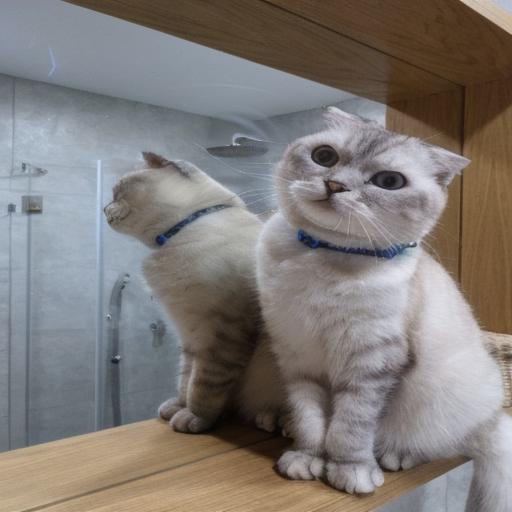}
    \caption{rec. ($\emptyset^A_t,\emptyset_t$)}
    \end{subfigure}
    \hfill
        \begin{subfigure}{0.24\linewidth} 
    \includegraphics[width=\linewidth]{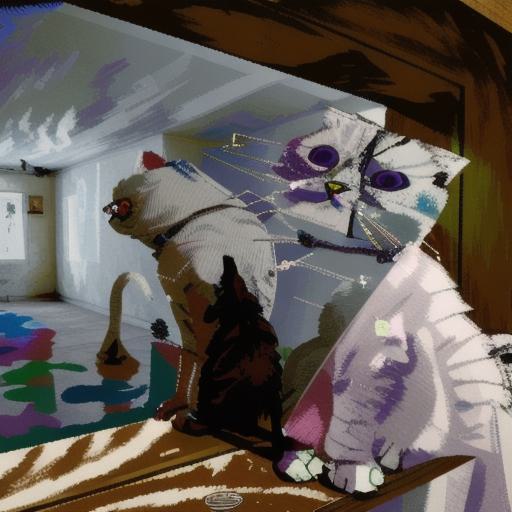}
    \caption{edit ($\emptyset^A_t,\emptyset_t$)}
    \end{subfigure}
    \hfill
    \begin{subfigure}{0.24\linewidth} 
    \includegraphics[width=\linewidth]{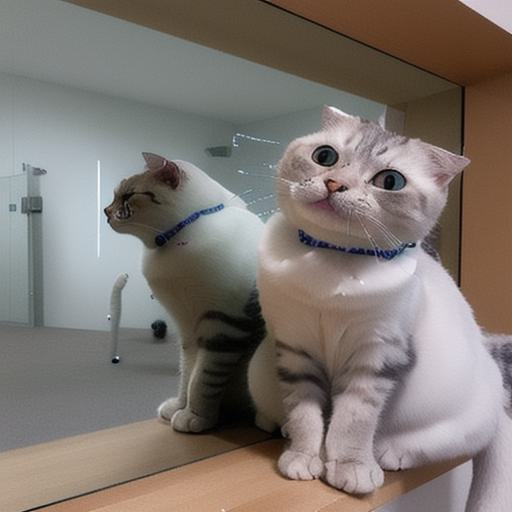}
    \caption{rec. ($\emptyset^A_t$)}
    \end{subfigure}
    \hfill
    \begin{subfigure}{0.24\linewidth} 
    \includegraphics[width=\linewidth]{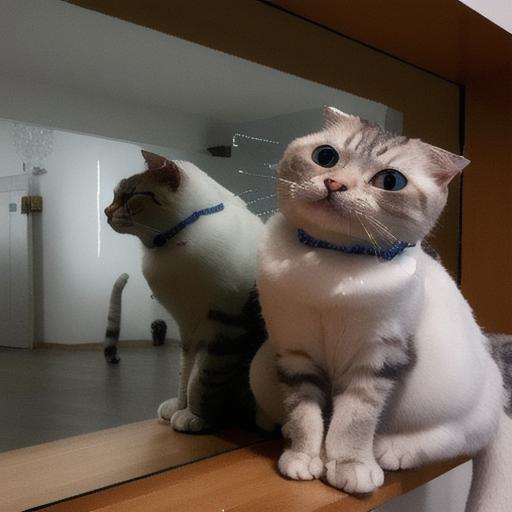}
    \caption{edit ($\emptyset^A_t$)}
    \end{subfigure}
    \caption{Reconstruction (rec.) and editing of blackpoint parameter of inverted real image. Editing using jointly optimized null conditionals ($\emptyset^A_t$,$\emptyset_t$) fails to adhere to edit direction, despite better initial reconstruction than only optimizing zero-param ($\emptyset^A_t$).}
    \label{fig:reconstrustrction_inversion}
\end{figure}

    \begin{figure*}[p]
        \centering
        \begin{subfigure}[b]{0.15\textwidth}
            \includegraphics[width=\textwidth]{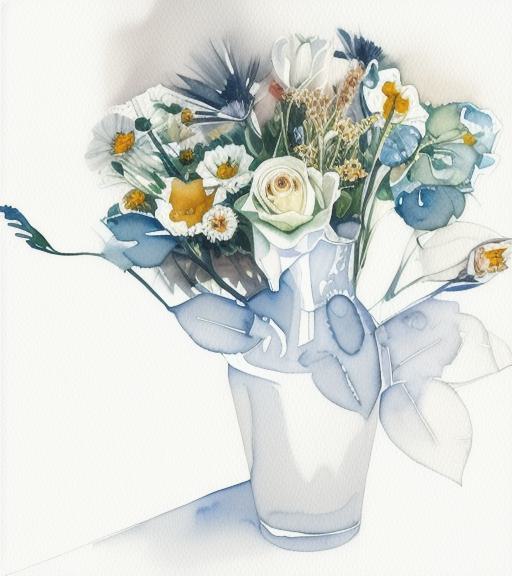}
            \caption{Input to ours}
            \label{fig:ours_a0.3_p0.15}
        \end{subfigure}
        \begin{subfigure}[b]{0.15\textwidth}
            \includegraphics[width=\textwidth]{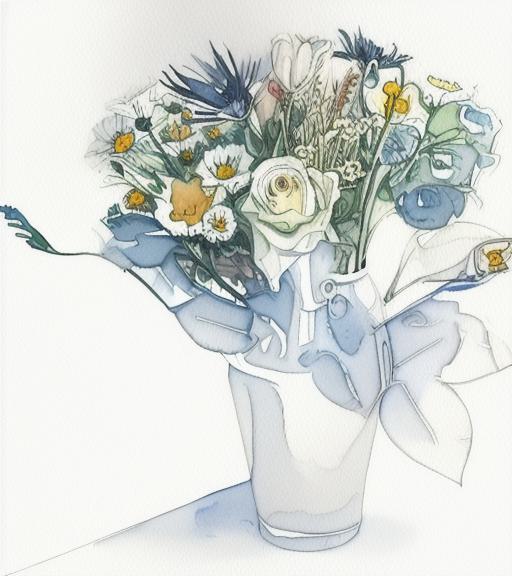}
            \caption{act=0.3, $\lambda=0.3$}
            \label{fig:ours_a0.3_p0.3}
        \end{subfigure}
        \begin{subfigure}[b]{0.15\textwidth}
            \includegraphics[width=\textwidth]{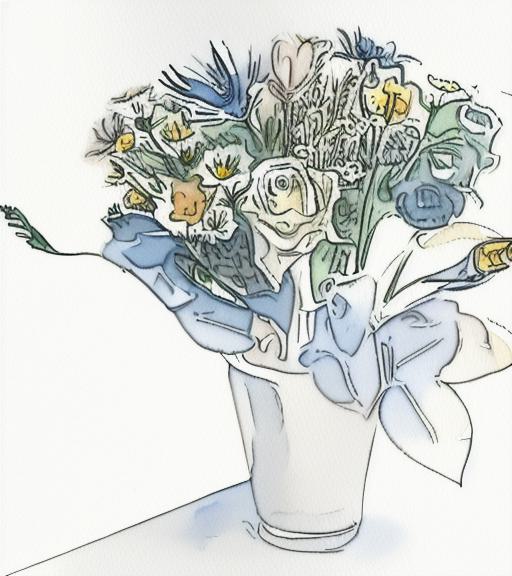}
            \caption{act=0.3, $\lambda=0.6$}
            \label{fig:ours_a0.3_p0.6}
        \end{subfigure}
        \begin{subfigure}[b]{0.15\textwidth}
            \includegraphics[width=\textwidth]{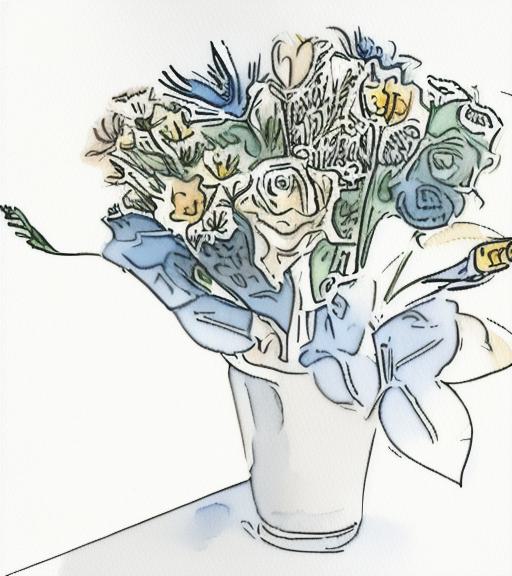}
            \caption{act=0.3, $\lambda=0.9$}
            \label{fig:ours_a0.3_p0.9}
        \end{subfigure}
        \begin{subfigure}[b]{0.15\textwidth}
            \includegraphics[width=\textwidth]{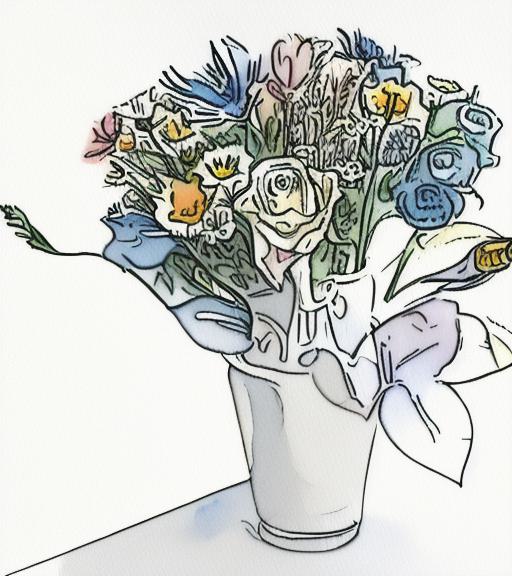}
            \caption{act=0.1, $\lambda=0.6$}
            \label{fig:ours_a0.1_p0.6}
        \end{subfigure}
        \begin{subfigure}[b]{0.15\textwidth}
            \includegraphics[width=\textwidth]{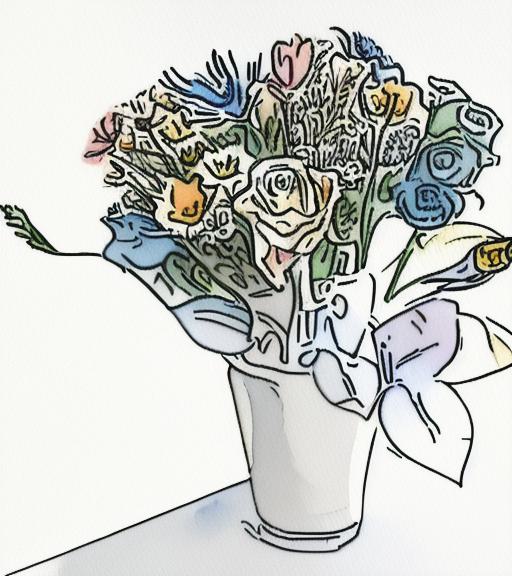}
            \caption{act=0.1, $\lambda=0.9$}
            \label{fig:ours_a0.1_p0.9}
        \end{subfigure}
        
        \begin{subfigure}[b]{0.15\textwidth}
            \includegraphics[width=\textwidth]{graphics/flower_contourWidth/input.jpg}
            \caption{Input to IP2P }
            \label{fig:input_ip2p}
        \end{subfigure}
        \begin{subfigure}[b]{0.15\textwidth}
            \includegraphics[width=\textwidth]{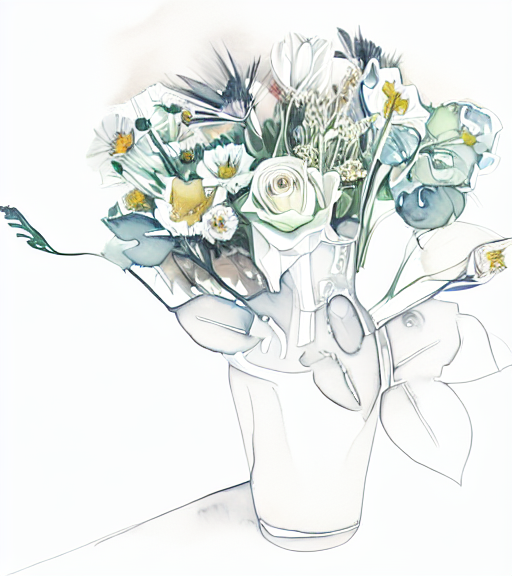}
            \caption{\footnotesize cfg=12}
            \label{fig:ip2p_cfg12}
        \end{subfigure}
        \begin{subfigure}[b]{0.15\textwidth}
            \includegraphics[width=\textwidth]{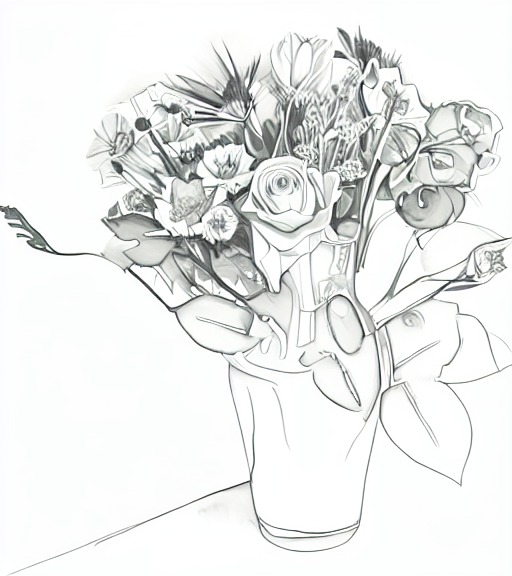}
            \caption{\footnotesize cfg=13}
            \label{fig:ip2p_cfg13}
        \end{subfigure}
        \begin{subfigure}[b]{0.15\textwidth}
            \includegraphics[width=\textwidth]{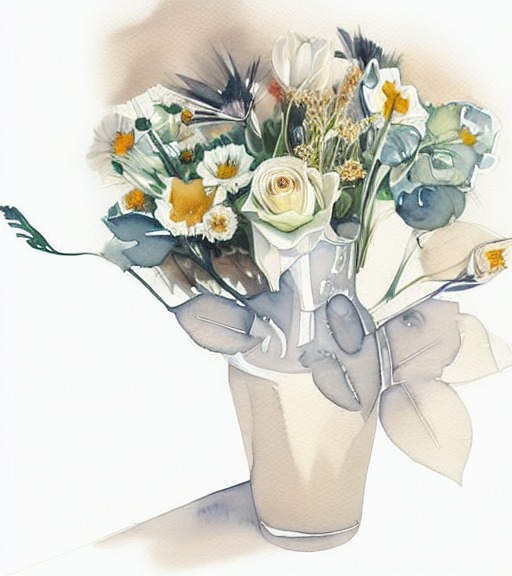}
            \caption{\footnotesize ``very'' + $P$}
            \label{fig:ip2p_thick_strokes}
        \end{subfigure}
        \begin{subfigure}[b]{0.15\textwidth}
            \includegraphics[width=\textwidth]{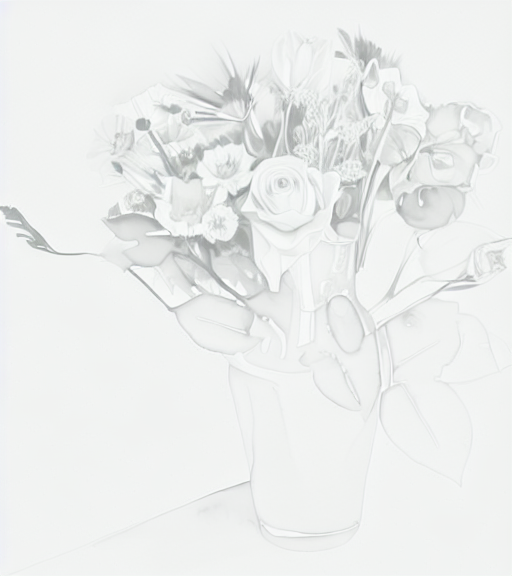}
            \caption{\footnotesize $P_{\text{outlines}}$}
            \label{fig:ip2p_add_outlines}
        \end{subfigure}
        \begin{subfigure}[b]{0.15\textwidth}
            \includegraphics[width=\textwidth]{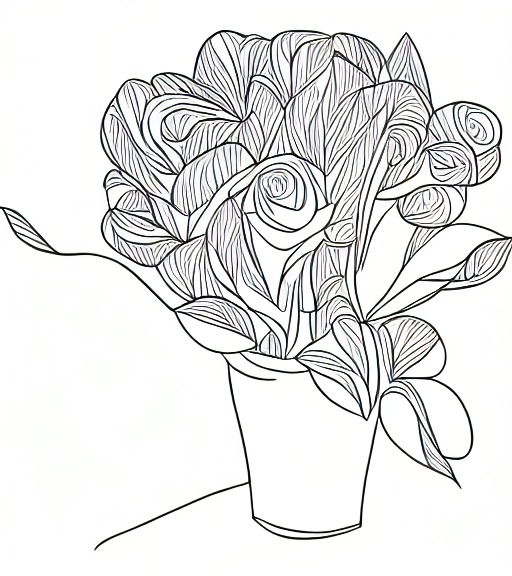}
            \caption{\footnotesize ``thick" + $P_{\text{outlines}}$}
            \label{fig:ip2p_thick_outlines}
        \end{subfigure}
        
        \caption{Comparison of continuous editing. Top row: our method varies contour width smoothly by adjusting parameter values and activation ($t$) time steps, demonstrating consistent control over line thickness. Bottom row: Using IP2P \cite{brooksInstructPix2PixLearningFollow2023} with same input, we attempt to replicate an increase in contour width using editing prompts or varying CFG. In (h) and (i) we apply the editing prompt $P$=``make it have thick strokes" and text cfg of 12/13, prepending a ``very" in (j). In (k) we use $P_{\text{outlines}}$ = ``Add outlines'', and in (l) $P_{\text{outlines}}$ = ``thick outlines''. }
        \label{fig:contour_width_variation}
        \vspace{0.5em}

        \includegraphics[trim={0cm 15.2cm 0cm 0cm}, width=0.97\linewidth, clip]{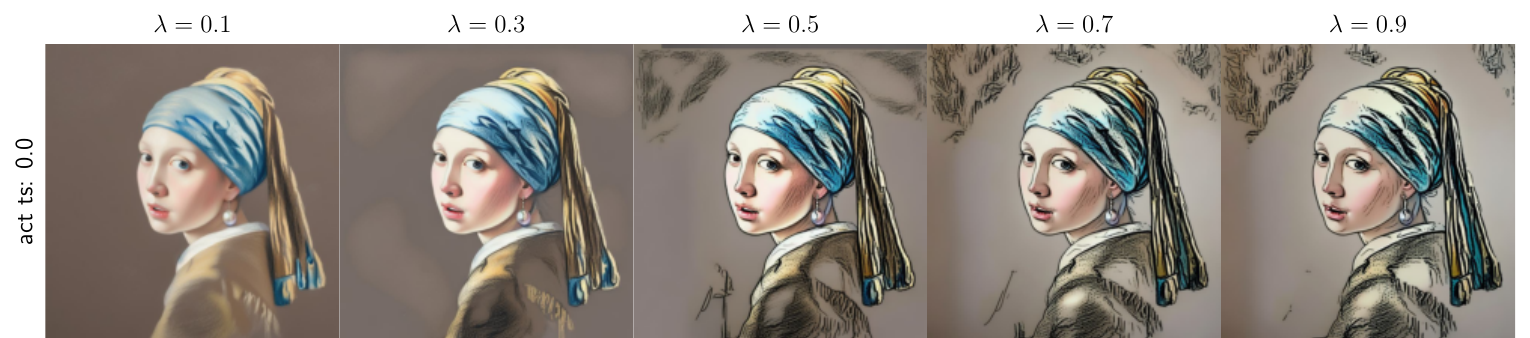}
        \includegraphics[trim={0cm 0cm 0cm 15.7cm}, width=0.97\linewidth, clip]{graphics/vary_act_ts/param_act_grid.pdf}
        \vspace{-1em}
        \caption{Varying the activation time step and the line-width ($\lambda$). Generation prompt: ``a pastel drawing of a girl with a pearl earring".}
        \label{fig:vary_act_p}

        \vspace{1em}
                
        \begin{subfigure}[b]{0.15\textwidth}
            \includegraphics[width=\textwidth,height=4.1cm]{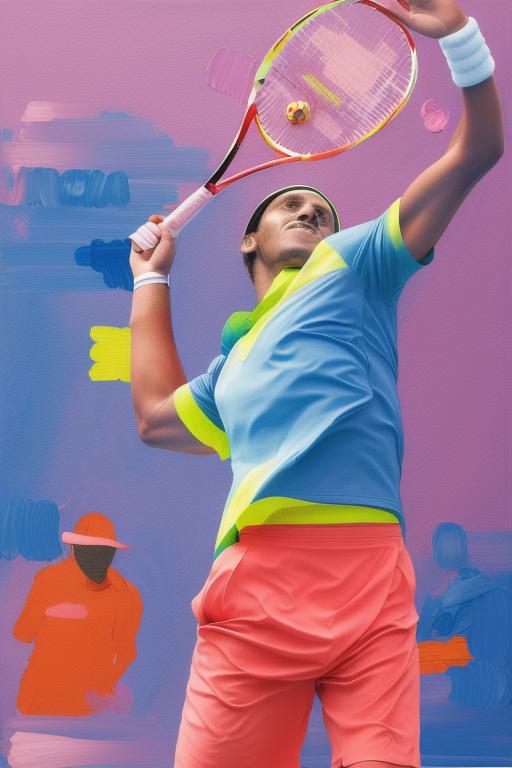}
            \caption{Input}
            \label{fig:tennis_p0.0}
        \end{subfigure}
        \begin{subfigure}[b]{0.15\textwidth}
            \includegraphics[width=\textwidth,height=4.1cm]{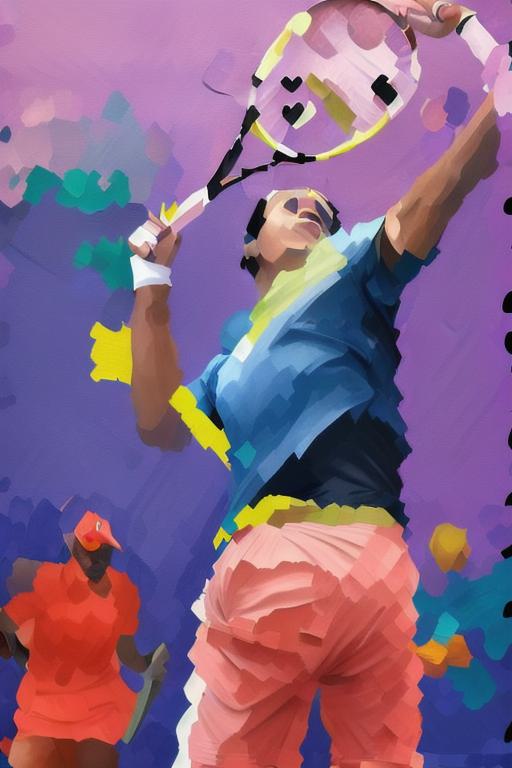}
            \caption{$\lambda=0.9$}
            \label{fig:tennis_p0.9}
        \end{subfigure}
        \begin{subfigure}[b]{0.15\textwidth}
            \includegraphics[width=\textwidth,height=4.1cm]{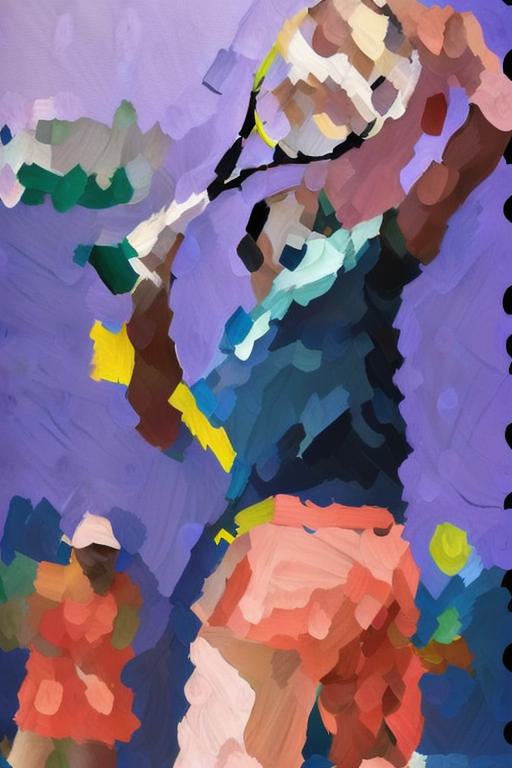}
            \caption{$\lambda=1.3$}
            \label{fig:tennis_p1.3}
        \end{subfigure}
        \begin{subfigure}[b]{0.15\textwidth}
            \includegraphics[width=\textwidth,height=4.1cm]{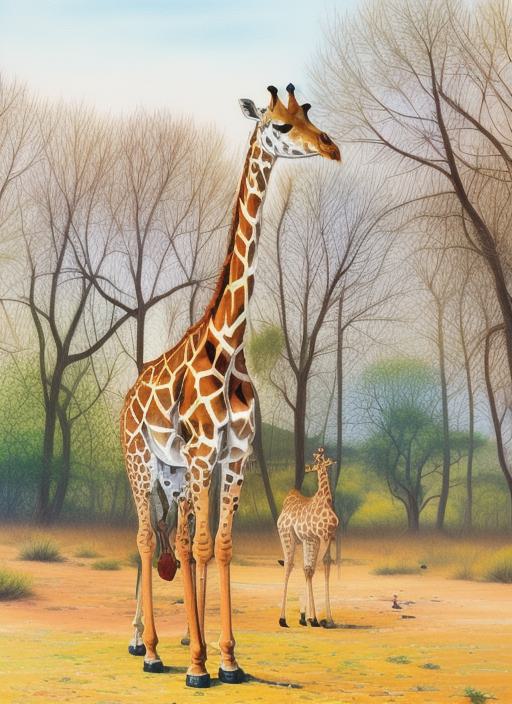}
            \caption{Input}
            \label{fig:giraffe_p0.0}
        \end{subfigure}
        \begin{subfigure}[b]{0.15\textwidth}
            \includegraphics[width=\textwidth,height=4.1cm]{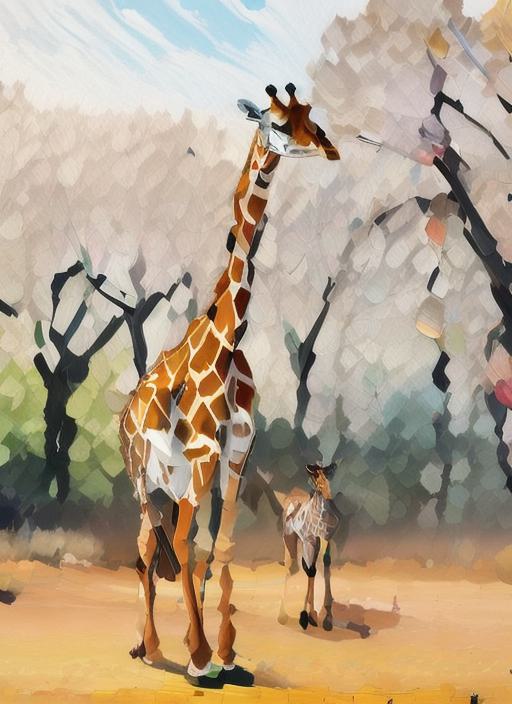}
            \caption{$\lambda=0.9$}
            \label{fig:giraffe_p0.9}
        \end{subfigure}
        \begin{subfigure}[b]{0.15\textwidth}
            \includegraphics[width=\textwidth,height=4.1cm]{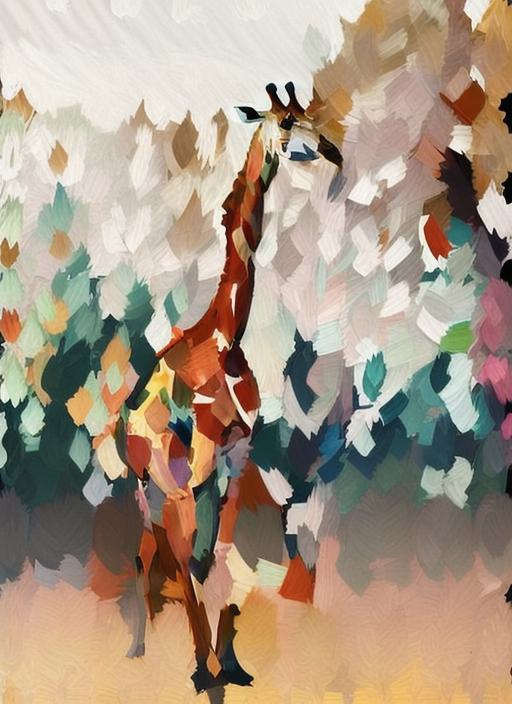}
            \caption{$\lambda=1.7$}
            \label{fig:giraffe_p1.7}
        \end{subfigure}
        \vspace{-0.5em}
        \caption{Stroke size variations using the PaintTransformer \cite{liu2021paint}-dataset finetuned model. Left prompt: "A colourful painting of a tennis player"; Right prompt: "A colourful painting of a giraffe". Please zoom in to compare details.}
        \label{fig:stroke_variation}
    \end{figure*}

\paragraph*{Null-conditional Guidance.}
In \cref{fig:reconstrustrction_inversion}, we ablate the null conditional guidance optimization by comparing optimization of only the null-parameter conditional, and joint optimizing both. While the joint optimization of ($\emptyset_t,\emptyset^A_t$) is able to more accurately reconstruct the input image, it fails to adhere to the parameter condition during editing and introduces significant structures. After optimizing solely $\emptyset^A_t$, the parameter can be edited in a stable manner.

\paragraph*{Activation Threshold.}  
The choice of activation threshold and parameter value significantly impacts the effect strength and image stability. 
\begin{wrapfigure}{r}{0.4\linewidth}
    \centering
    \includegraphics[width=\linewidth]{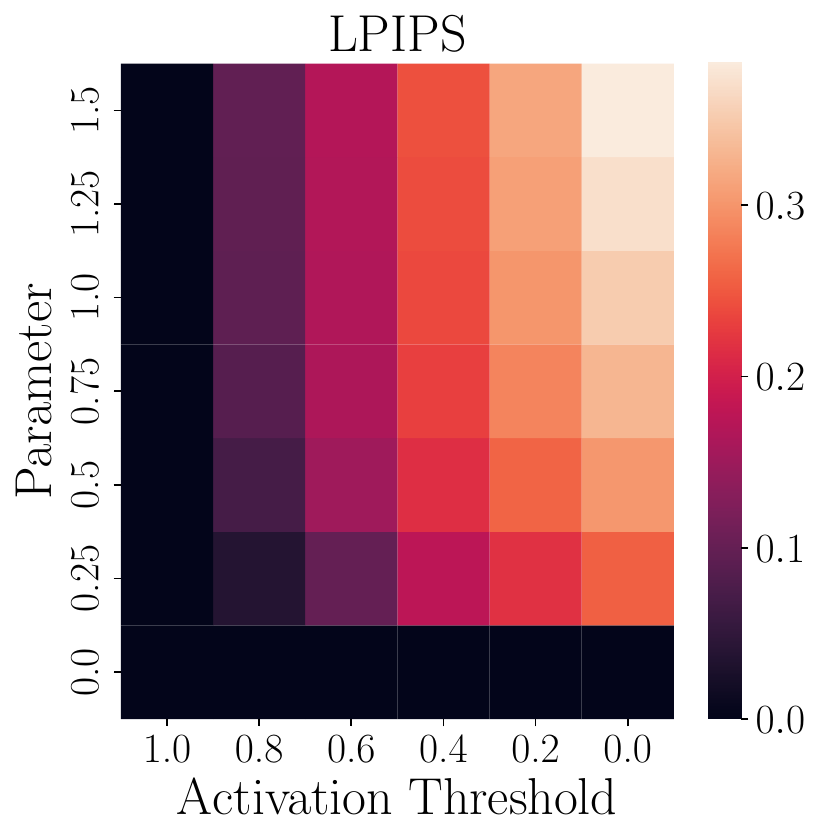}
    \caption{Change in output compared to original $\varepsilon_\theta$ only.}
    \label{fig:metric_heatmap}
\end{wrapfigure}
To evaluate these influences, we vary the outline thickness and the activation timesteps $\text{act}_t$ when generating $D_T$. \cref{fig:metric_heatmap} shows a heatmap of the resulting changes in \ac{LPIPS} compared to the original ControlNet-generated images (i.e., where $w_2$=0), highlighting how both factors contribute to the magnitude of stylistic alterations.

%% file: results.tex
\begin{figure*}[t]
    \centering
    \includegraphics[trim={0.2cm 22.7cm 0cm 0cm},clip, width=0.8\linewidth]{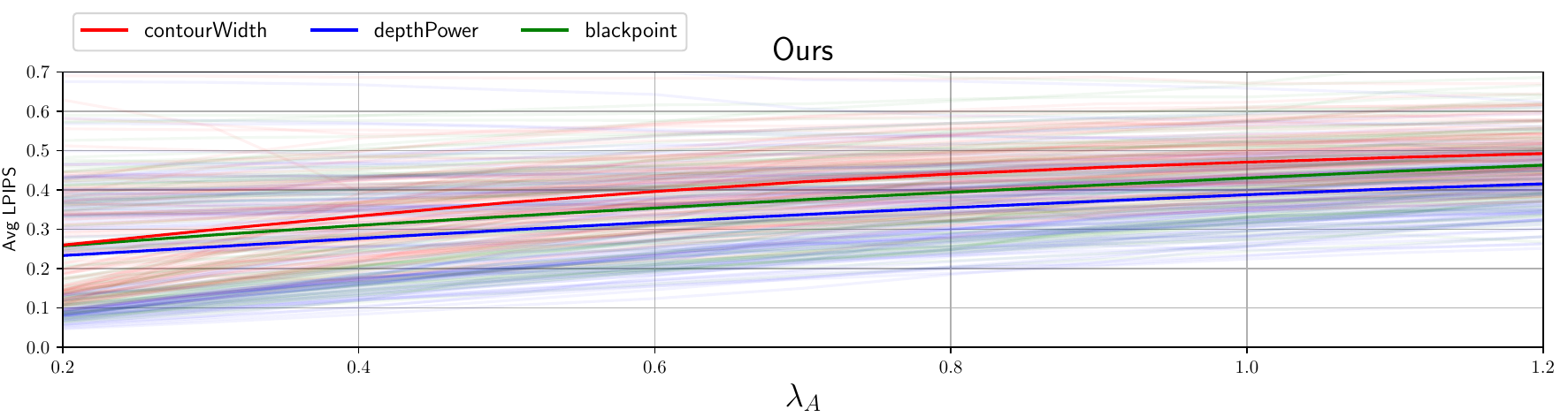}\\
    \includegraphics[trim={0.2cm 7.5cm 0cm 15.7cm},clip, width=0.8\linewidth]{graphics/plots/parameter_cfg_influence_lpips.pdf}
    \caption{
    Influence of increasing stylization strength. LPIPS against the unedited inputs are plotted, each thin line represents an individual image from our testing dataset edited by varying parameter strengths, for three parameters. Opaque lines denote the mean per parameter. For our approach, the parameter ($\lambda_A$) is incrementally adjusted, while InstructPix2Pix~\cite{brooksInstructPix2PixLearningFollow2023} uses the classifier-free guidance scale. %
    }
    \label{fig:parameter_cfg_influence}
\end{figure*}

\begin{figure}
    \centering
        \begin{subfigure}{0.485\linewidth}
            \includegraphics[width=\linewidth]{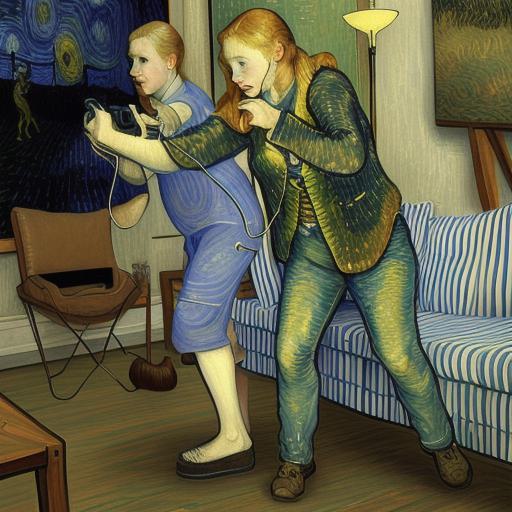}
            \caption{Input}
        \end{subfigure}
        \begin{subfigure}{0.485\linewidth}
            \includegraphics[width=\linewidth]{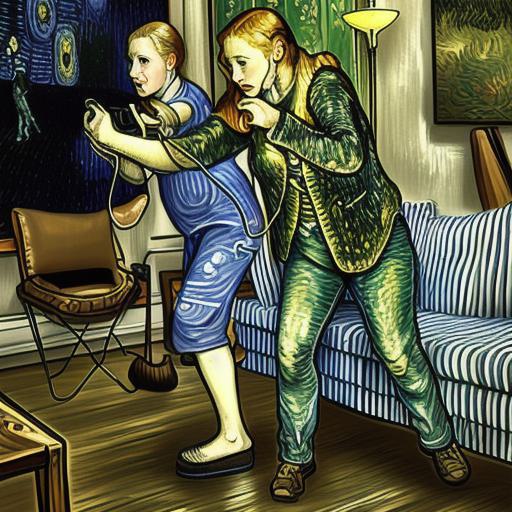} 
            \caption{Ours}
        \end{subfigure}\\
        \begin{subfigure}{0.485\linewidth}
            \includegraphics[width=\linewidth]{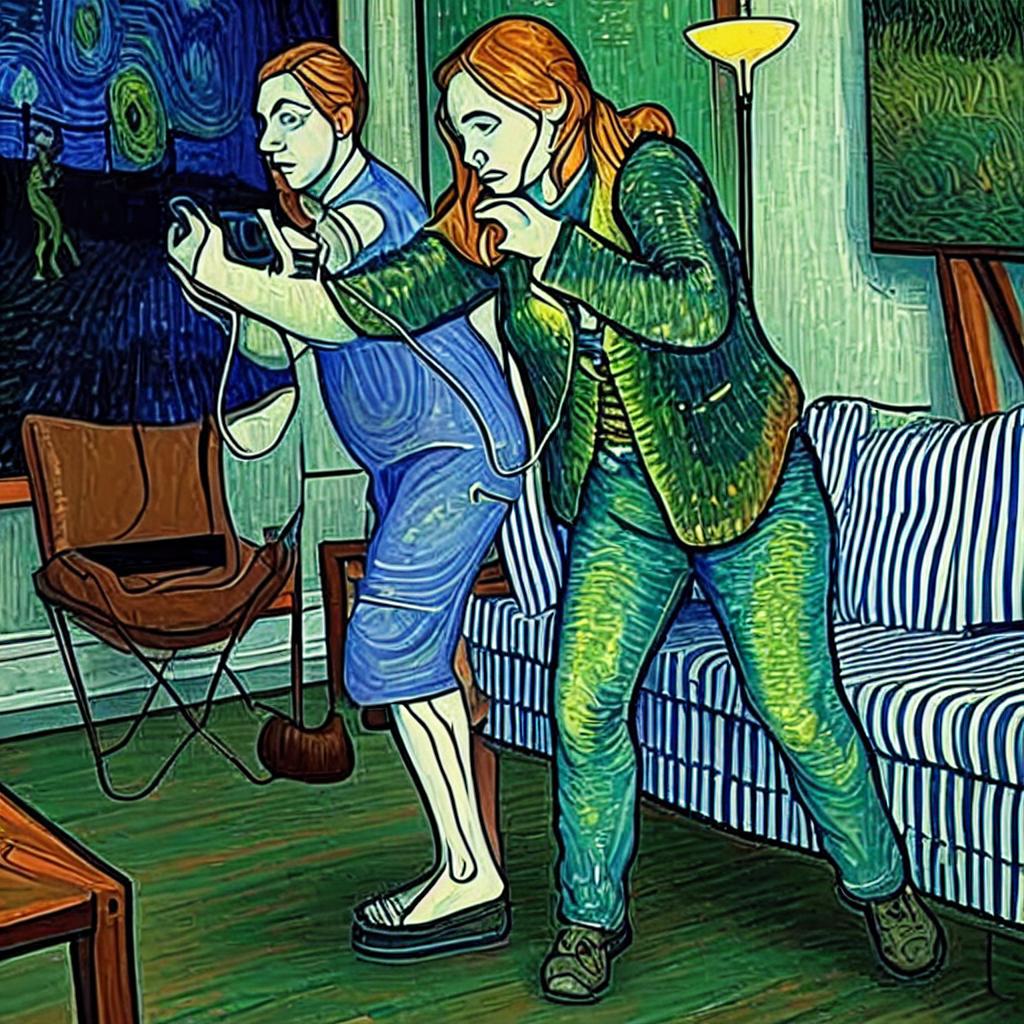}
            \caption{StyleBooth \cite{han2024stylebooth}}
        \end{subfigure}
        \begin{subfigure}{0.485\linewidth}
            \includegraphics[width=\linewidth]{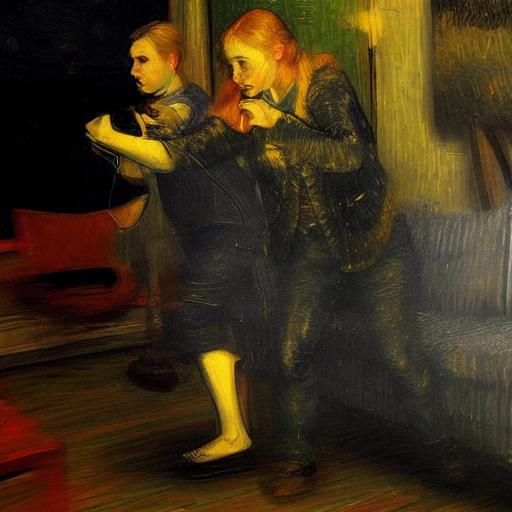}
            \caption{InstructPix2Pix \cite{brooksInstructPix2PixLearningFollow2023}}
            \label{fig:examplesUserStudy:ip2p}
        \end{subfigure}
        \caption{Sample images from user study. Here the users where asked to judge which image better captures the editing instruction of ``Increase the local contrast and highlights". }
        \label{fig:examplesUserStudy}
        \vspace{-1em}
\end{figure}

\section{Results}

\subsection{Qualitative}
In \cref{fig:contour_width_variation} we vary the contour width parameter of our approach and compare to attempts of trying to achieve a similar effect with InstructPix2Pix (IP2P)  \cite{brooksInstructPix2PixLearningFollow2023}. We can smoothly vary the line width using our approach. However for IP2P, we were not able to increase the stroke size without severly altering the image, despite trying a various combinations of text prompts, and text and image CfG scales. We also noticed that even small changes in parameter space often lead to abrupt changes and inconsistencies. In \cref{fig:stroke_variation} we show results of using the PaintTransformer-tuned model \cite{liu2021paint} to introduce stroke patches into images. A higher size parameter yields increasingly rough patches, that are deeply integrated as paint strokes into the image. We also use the PaintTransformer model in \cref{fig:teaser} (bottom row, left), where it seamlessly integrates into the watercolor style and increases the splattering effect.

\begin{table}
\centering
\caption{User study results. Participants were asked to select images that adhere best both to the original image and to the editing prompt, describing an image edit.}
\label{tab:methods_by_param}
\begin{tabular}{lcccc}
\toprule
Editing Task & IP2P & InstrDiff & StyleBooth & Ours \\
\midrule
Contours  & \underline{27\%} & 2\% & 14\% & \textbf{57\%} \\
Black point  & \textbf{41\%} & 16\% & 10\% & \underline{33\%} \\
Local contrast  & \underline{24\%} & 2\% & 16\% & \textbf{58\%} \\
\midrule
Total & \underline{30\%} & 7\% & 13\% & \textbf{50\%} \\
\bottomrule
\end{tabular}
\end{table}

\paragraph*{User Study.} We conducted a user study comparing our method with instruction-based \ac{SD} editing approaches, as no purely parametric alternatives are available. Specifically, we evaluated against InstructPix2Pix \cite{brooksInstructPix2PixLearningFollow2023}, InstructDiffusion (InstrDiff) \cite{Geng23instructdiff}, and StyleBooth \cite{han2024stylebooth}. Using our test dataset (\cref{subsec:ablationstudies}), which includes 50 images generated with photographic and Van Gogh prompts, we produced results with each method for three different parameters by translating these parameters into prompts, trying out several variants and CfG settings and choosing those that best reflect our visual attribute dataset e.g., depth power parameter as ``make it have an intense depth contrast and highlights."
Participants were provided a brief explanation of the tasks, focusing on adherence to the editing prompt and maintaining the style and content of the original image. They were then presented with the original and four edited versions arranged in random order, asked to select their preferred result. We gathered 300 task samples from 50 participants revia \href{prolific.com}{prolific.com}, with four responses per task. To filter ambiguous results, we retained tasks where at least three participants agreed on the preferred method, resulting in 235 usable samples.
The results, shown in \cref{tab:methods_by_param}, indicate that our method is generally preferred over the other approaches. In some cases, InstructPix2Pix produced more stable results for simpler tasks, such as darkening specific areas. We show a comparison for depth contrast in \cref{fig:examplesUserStudy}, excluding InstrDiff\cite{Geng23instructdiff} as its output failed to preserve the semantic content. More visual examples, details of the study and a screenshot of the study interface are provided in the supplemental material.

\subsection{Quantitative}
\label{subsec:quantative}

\begin{table}[t]
\centering
\caption{Similarity values to originals in test dataset. We compute the \ac{LPIPS} \cite{zhang2018perceptual}, CLIPScore between images \cite{radford2021learningCLIP} and NST style loss \cite{gatys2016image}. Note that these values have to be interpreted with nuance (lower/higher $\neq$ better), see \cref{subsec:quantative}.}
\label{tab:metrics_relatedmethods}
\begin{tabular}{lccc}
\toprule
\textbf{Approach}            & \textbf{LPIPS}$\downarrow$ & $\textbf{CLIP}_{\text{img}}\uparrow$ & \textbf{Style}$\downarrow$ \\
\midrule
Null Text Invers. \cite{mokadyNULLTextInversionEditing}          & 0.452              & 0.793             & $<$0.001              \\
Instruct Pix2Pix \cite{brooksInstructPix2PixLearningFollow2023}            & 0.259              & 0.907             & 0.007              \\
MagicBrush \cite{zhang2024magicbrush}                  & 0.123              & 0.964             & 0.001              \\
InstructDiffusion \cite{Geng23instructdiff}           & 0.467              & 0.820             & 0.008              \\
Stylebooth \cite{han2024stylebooth}                  & 0.597              & 0.817             & 0.004              \\ 
Ours                           & 0.343              & 0.894             & 0.001             \\
\bottomrule
\end{tabular}
\end{table}

\paragraph*{Similarity metrics.} In \cref{tab:metrics_relatedmethods} we show similarity metrics in terms of perceptual (LPIPS), semantic (CLIP) and stylistic~(NST~\cite{gatys2016image} style loss) similarity of editing methods to the original on our test dataset with the same editing captions as above.
Following human–calibrated thresholds \cite{zhang2018perceptual}, LPIPS $<\!0.15$ is typically imperceptible, whereas values $>\!0.45$ denote major content change.
We compute CLIP$_{\text{img}}$, the cosine similarity of OpenCLIP image embeddings \cite{radford2021learningCLIP}, which is found to capture human judgment of semantic scene similarity well ($\sim 87\%$ human alignment for NIGHT~\cite{fu2023dreamsim} for OpenCLIP). From experimentation, values \mbox{CLIP\textsubscript{img}$>0.92$} generally correspond to \emph{near-identity} semantics, whereas the range \mbox{$0.80\!-\!0.92$} would correspond to faithful but visibly edited images, and \mbox{$<0.80$} to a region where semantic drift is strong.

MagicBrush~\cite{zhang2024magicbrush} attains the highest CLIP$_{\text{img}}$ but the lowest LPIPS, confirming that it largely leaves the image unchanged despite the prompt.
Conversely, null-text inversion~\cite{mokadyNULLTextInversionEditing} shows the largest LPIPS and lowest CLIP$_{\text{img}}$ evidencing over-editing artefacts.
InstructPix2Pix~\cite{brooksInstructPix2PixLearningFollow2023} strikes a better trade-off but introduces the greatest stylistic drift, higher style scores typically indicate change in local textures and color - which would not be expected from the edits in our test dataset (e.g., \cref{fig:examplesUserStudy:ip2p}).
Our method yields a mid-high CLIP$_{\text{img}}$ while keeping LPIPS and style loss 
 inside the “visible-but-faithful’’ band, indicating salient yet well-integrated fine-stylistic edits that preserve the original semantics.

\begin{table}[]
    \centering
    \begin{tabular}{lccc}
    \toprule
        Method & contour-width & depth-power & blackpoint \\
    \midrule
        InstructPix2Pix & 33.4 &  49.7 & 51\\
        InstructDiffusion & 32.1 & 66.6 & 48.2\\
        StyleBooth & 16.6 & 18.5 & 17.5 \\
        Ours & \textbf{9.3} & \textbf{14.9} & \textbf{13.7}\\
    \bottomrule
    \end{tabular}
    \caption{Smoothness of change measured in ISTD~\cite{guo2024smooth} on the tasks in \Cref{fig:parameter_cfg_influence}. }
    \label{tab:placeholder}
\end{table}

\paragraph*{Continuous control.} We compare the behaviour of state-of-the-art methods in adjusting stylization strength. Unlike our method, these methods approaches lack an explicit control parameter for stylization strength; instead, we adjust the text guidance scale of the editing prompt to approximate similar behavior. In \cref{fig:parameter_cfg_influence}, we visualize the impact of increasing stylization strength by plotting LPIPS values against the unedited input images.
Given the high diversity of samples, averaging LPIPS alone does not capture meaningful variations; hence, each thin line represents an individual input image incrementally stylized by increasing a single parameter’s strength. The graph demonstrates that our approach provides more consistent and predictable editing behavior compared to InstructPix2Pix, which displays abrupt jumps in LPIPS, making fine-grained stylization adjustments infeasible. Quantitatively, we measure smoothness using the ISTD metric~\cite{guo2024smooth} over the same set of generated images (increasing lambda or CfG). Compared to the methods in the user study, our approach significantly improves the ISTD.  See supplemental for the plots for these methods.  

%% file: conclusions.tex
\section{Discussion}

We presented a method for precise control of stylistic attributes in \acp{LDM} using stylistic control networks. By finetuning on datasets with stylistic effects applied at varying strengths, our approach learns disentangled editing directions mapped to explicit artistic control parameters. The guidance mechanism selectively amplifies the desired edits while suppressing irrelevant information and integrates seamlessly with the visual characteristics of the base Stable Diffusion model. This learning framework can reasonably be applied to any visual attribute parameterizable through a control setting.

\paragraph*{Limitations} The stylistic control network adds runtime and memory consumption. Adding new stylistic controls requires additional training data, and parameter adjustments for strength and activation timesteps can lead to trade-offs between effect intensity and content preservation. Additionally, real image inversion and subsequent editing can introduce artifacts due to suboptimal inversions.

Our approach represents a step towards bridging the gap between traditional image editing and generative techniques. We believe it offers a promising direction for enhancing user control in image synthesis, providing a foundation for future research to further refine and expand the capabilities of LDMs in artistic and professional workflows.

%% file: supplemental_content.tex
\section{Background}

Diffusion models progressively add random noise to input data through a variance-preserving Markov process~\cite{sohl2015deep,ho2020denoising,song2020score}. They then learn to reverse this process by denoising, thereby generating the desired data samples. Latent Diffusion Models~\cite{rombach2022high} apply this process in a low-dimensional latent space $z$, where a Variational Autoencoder (VAE)~\cite{kingma2013auto} first encodes an image $I$ into $z_0 = \mathcal{E}(I)$ using a pretrained encoder $\mathcal{E}(\cdot)$ and then noise is added to $z_0$ for $t=1...T$ timesteps until $z_T\sim \mathcal{N}(0,I)$.
During the denoising phase, the model predicts the noise delta $\varepsilon_{\theta}(z_t,t,c)$ at each timestep $t$, moving from $z_t$ to $z_{t-1}$. Here, $\varepsilon_{\theta}$ is a neural network, often a U-Net~\cite{ronneberger2015u}, and $c$ represents conditioning information, in our case text embeddings $c_{\text{p}}$. The training loss in text-to-image models \cite{rombach2022high} minimizes the difference between the actual noise $\varepsilon \sim \mathcal{N}(0,I)$ and the predicted noise $\varepsilon_{\theta}$:
\begin{equation}
\label{eq:diffusion_loss}
L = \mathbb{E}_{\mathcal{E}(I), c_{\text{p}},\varepsilon,t}\left[\lVert \varepsilon-\varepsilon_{\theta}(z_t, t, c_{\text{p}}) \rVert_{2}^{2} \right], t=1,...,T
\end{equation}
After the model is trained, it can effectively denoise from $z_T$ back to $z_0$ using an efficient diffusion sampler~\cite{song2020denoising, lu2022dpm}.

Finally, the decoded image $I$ is reconstructed using a frozen decoder $\mathcal{D}(\cdot)$.
The model is trained concurrently on both conditional ($c$) and unconditional (null-text embedding $\emptyset$) objectives. During inference, the difference between unconditioned and (text) conditioned outputs, denoted as vector $\Vec{g_p}$, can guide the generative process towards the conditioned goal. This classifier-free guidance (CfG)~\cite{hojonathanClassifierFreeDiffusionGuidance2022} is computed as
\begin{align}
\Vec{g_p} &= \varepsilon_{\theta}(z_t,t, c_{\text{p}}) - \varepsilon_{\theta}(z_t,t,\emptyset) \\
 \varepsilon_{\theta}(z_t,t,g_p) &= \varepsilon_{\theta}(z_t,t,\emptyset) + wg_p
\label{eq:cfg}
\end{align}
where, $w$ is the CfG scale that adjusts the intensity of the guidance.

\begin{figure}[b]
    \centering
    \includegraphics[width=\linewidth]{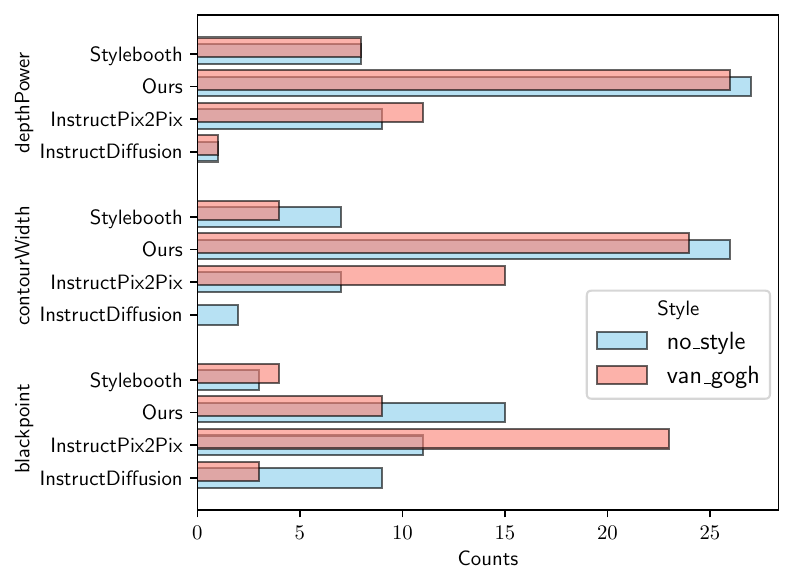}
    \caption{User method preferences. Shown are the preference counts of methods in editing tasks in which the method was chosen by the majority (at least 3 of 4 participants). We show results per-parameter and style variant (``no\_style" is the photography style). }
    \label{fig:userstudy_details}
\end{figure}

\section{User Study Details}
\label{sec:userstudydetails}
\begin{figure}[h]
    \centering
    \includegraphics[width=\linewidth]{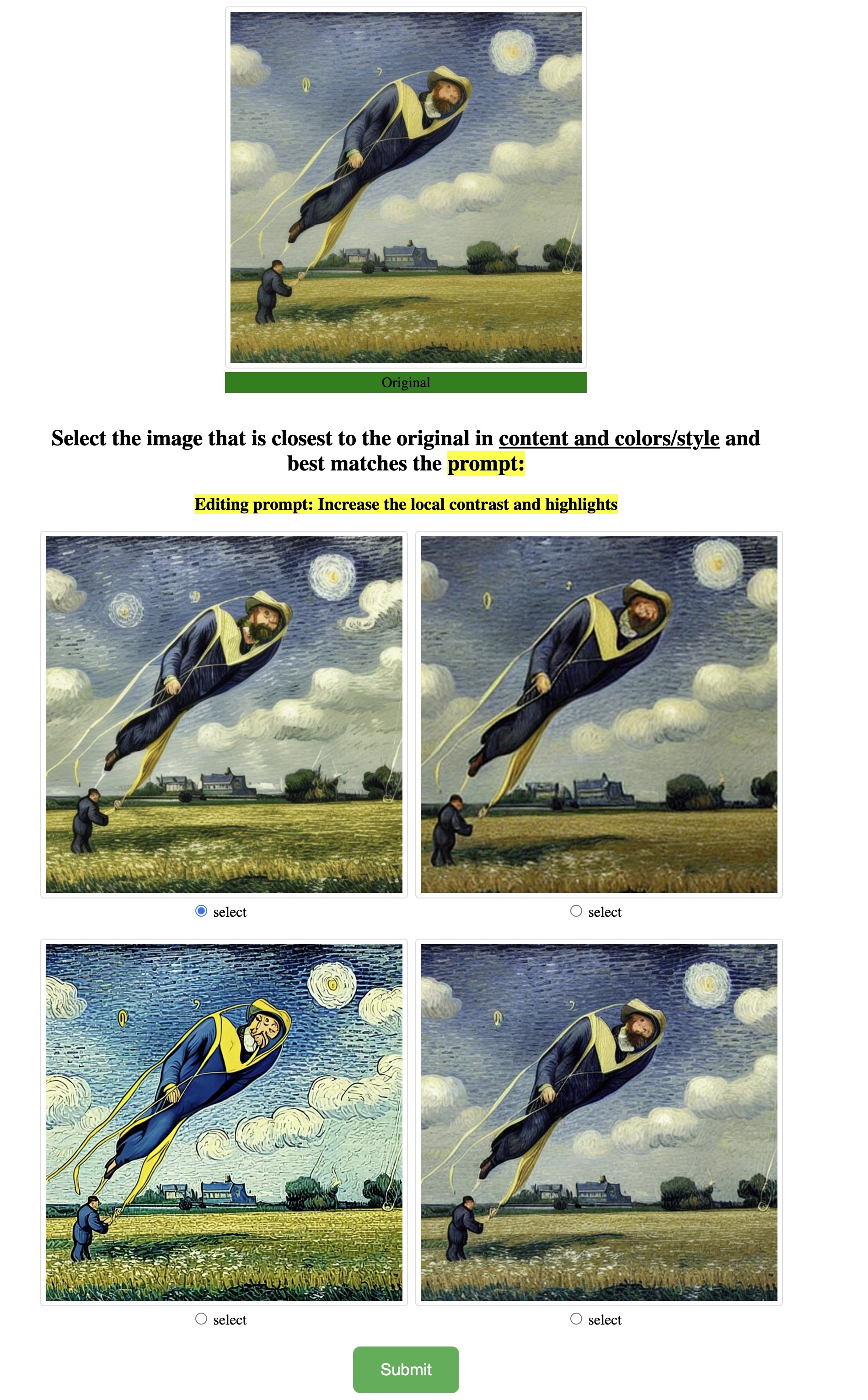}
    \caption{Screenshot from user the study}
    \label{fig:user_study_screenshot}
\end{figure}

\begin{figure*}
    \centering
    \begin{minipage}[c]{0.02\textwidth}
        \rotatebox[origin=c]{90}{Increase contours}
    \end{minipage}%
    \begin{minipage}[c]{\textwidth}
        \begin{subfigure}{0.19\textwidth}
            \caption*{Input}
            \includegraphics[width=\linewidth]{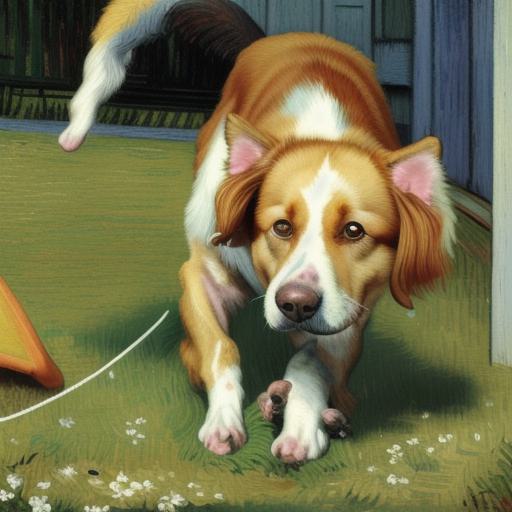}
        \end{subfigure}
        \begin{subfigure}{0.19\textwidth}
            \caption*{Ours}        
            \includegraphics[width=\linewidth]{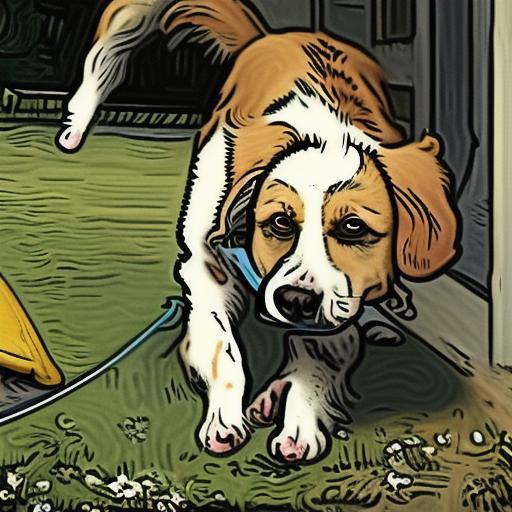}
        \end{subfigure}
        \begin{subfigure}{0.19\textwidth}
            \caption*{StyleBooth \cite{han2024stylebooth}}
            \includegraphics[width=\linewidth]{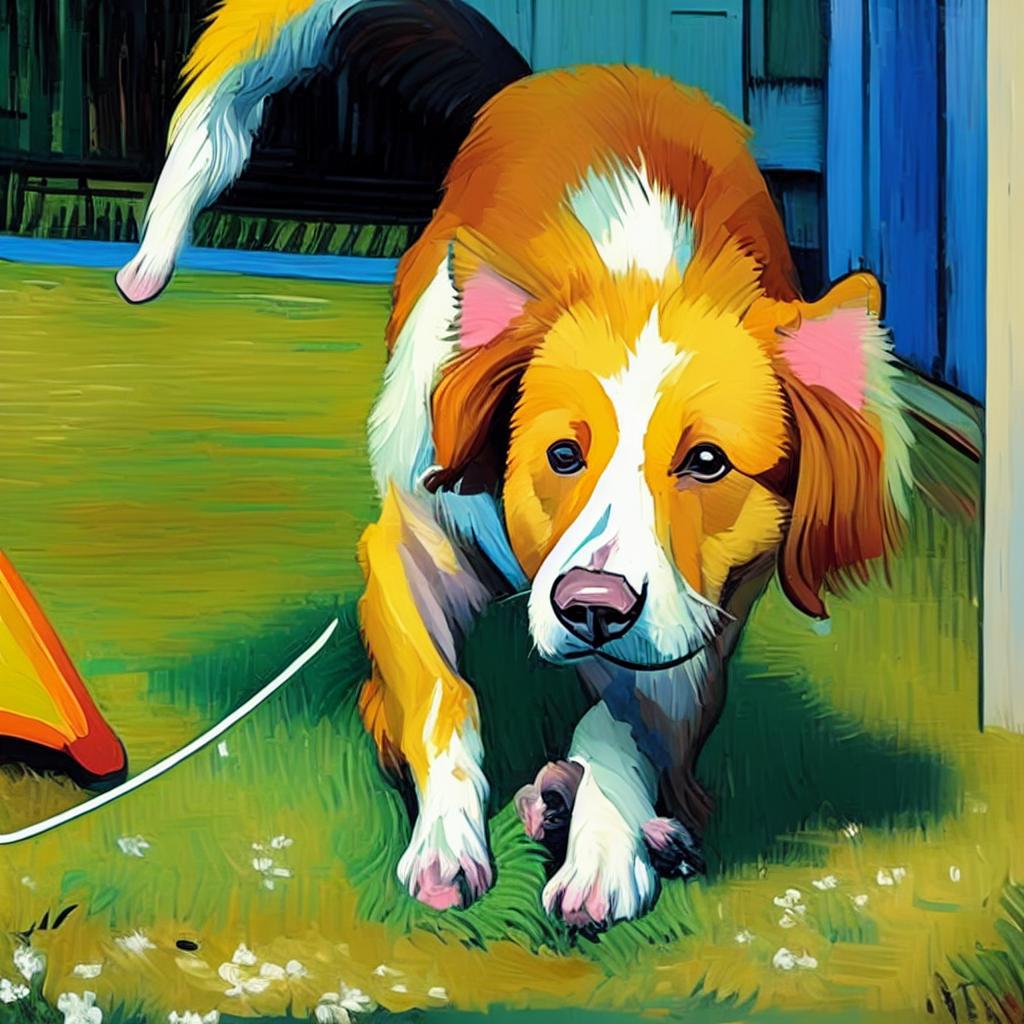}
        \end{subfigure}
        \begin{subfigure}{0.19\textwidth}
            \caption*{InstructDiffusion \cite{Geng23instructdiff}}
            \includegraphics[width=\linewidth]{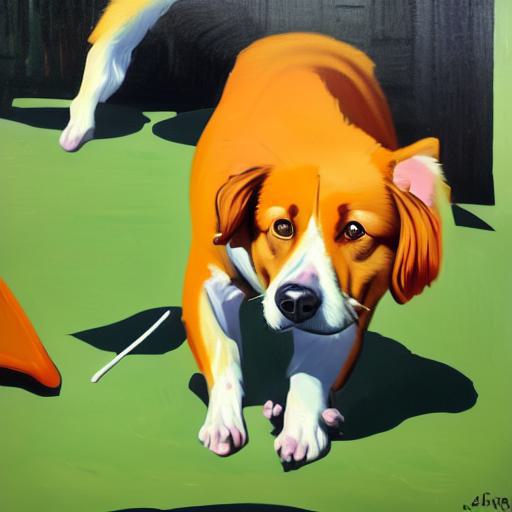}
        \end{subfigure}
            \begin{subfigure}{0.19\textwidth}
            \caption*{InstructPix2Pix \cite{brooksInstructPix2PixLearningFollow2023}}
            \includegraphics[width=\linewidth]{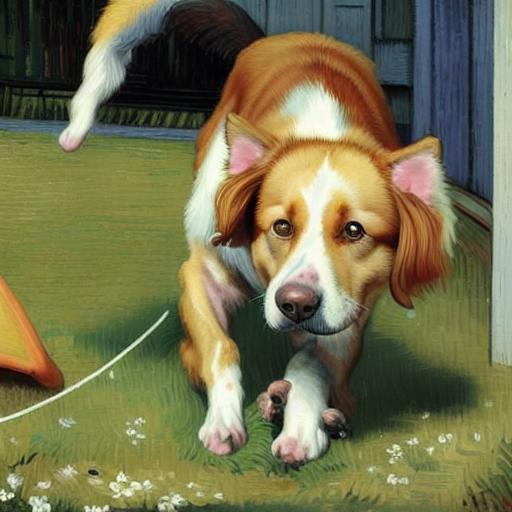}
        \end{subfigure}
        
        \begin{subfigure}{0.19\textwidth}
            \includegraphics[width=\linewidth]{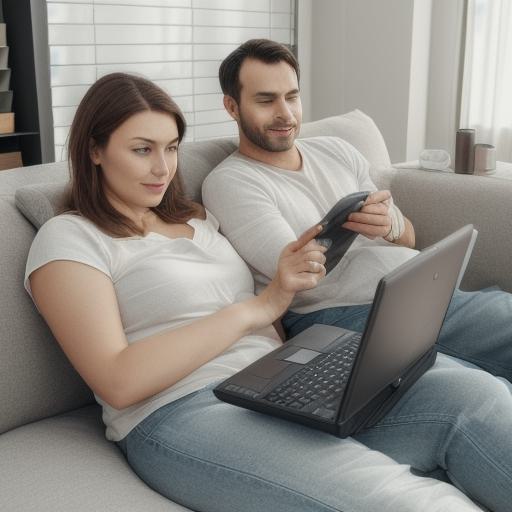}
        \end{subfigure}
        \begin{subfigure}{0.19\textwidth}
            \includegraphics[width=\linewidth]{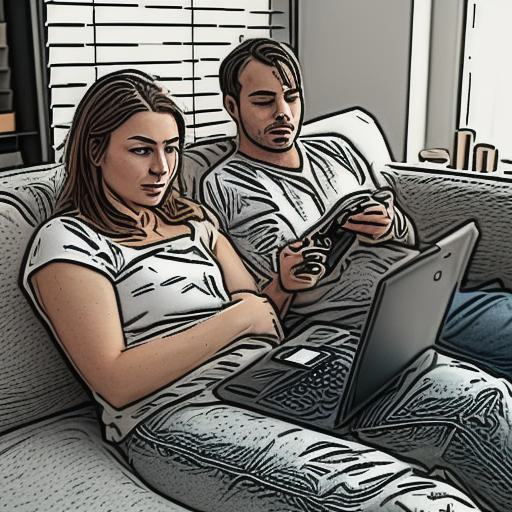}
        \end{subfigure}
        \begin{subfigure}{0.19\textwidth}
            \includegraphics[width=\linewidth]{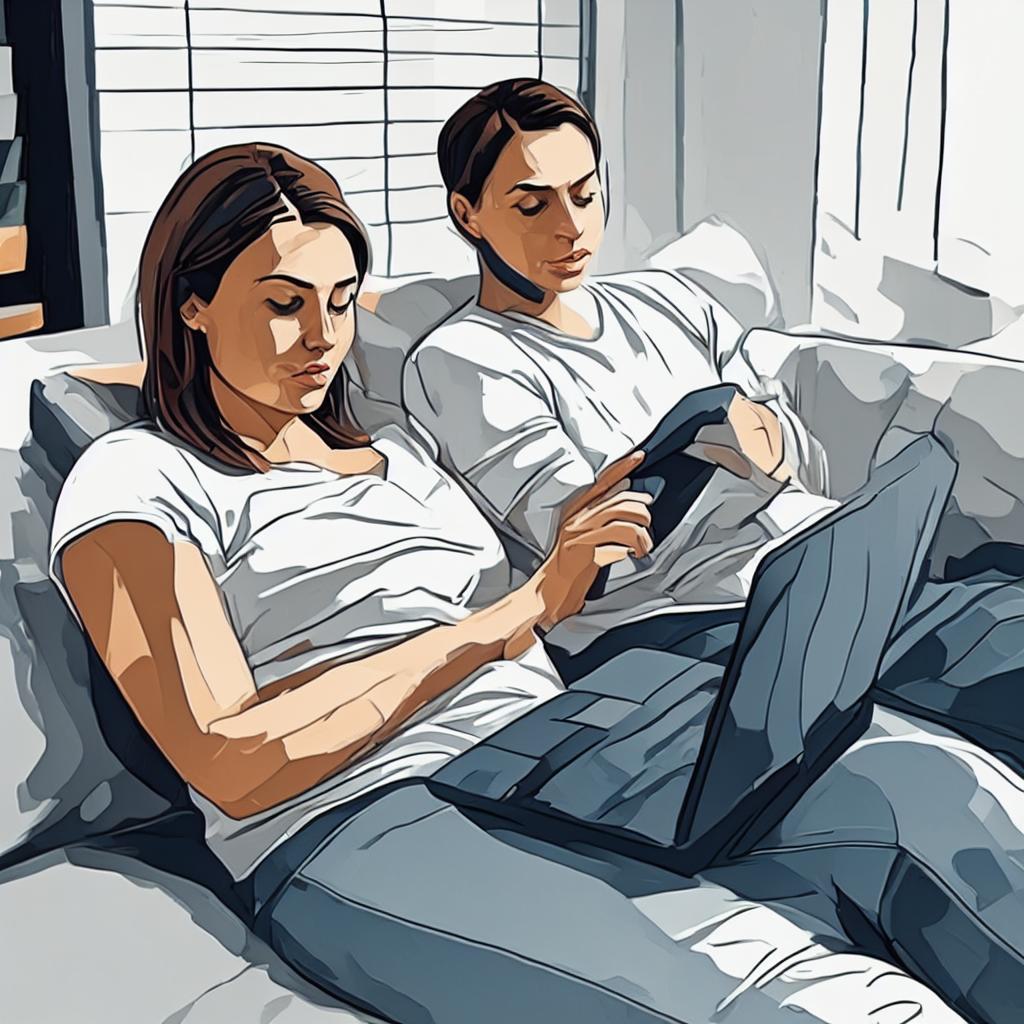}
        \end{subfigure}
        \begin{subfigure}{0.19\textwidth}
            \includegraphics[width=\linewidth]{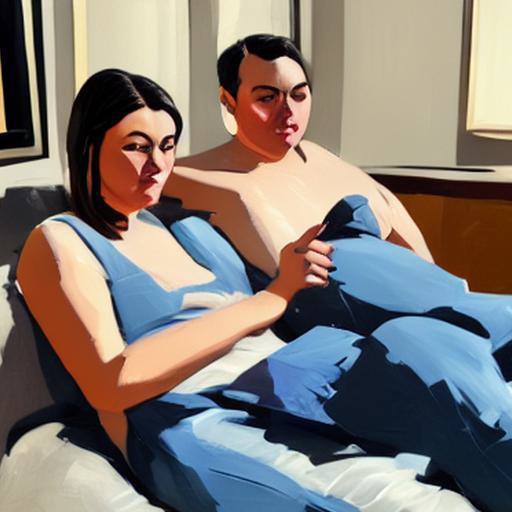}
        \end{subfigure}
            \begin{subfigure}{0.19\textwidth}
            \includegraphics[width=\linewidth]{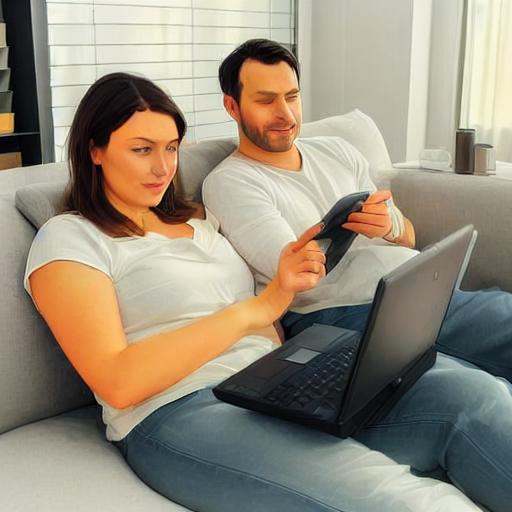}
        \end{subfigure}
    \end{minipage}\\[0.3em]

    \begin{minipage}[c]{0.02\textwidth}
        \rotatebox[origin=c]{90}{Lower blackpoint}
    \end{minipage}%
    \begin{minipage}[c]{\textwidth}
        \begin{subfigure}{0.19\textwidth}
            \includegraphics[width=\linewidth]{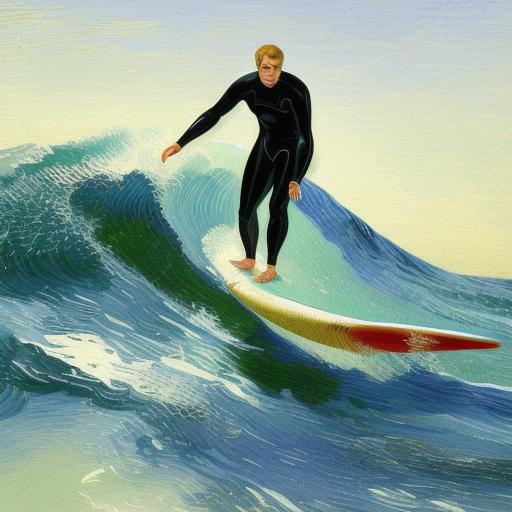}
        \end{subfigure}
        \begin{subfigure}{0.19\textwidth}
            \includegraphics[width=\linewidth]{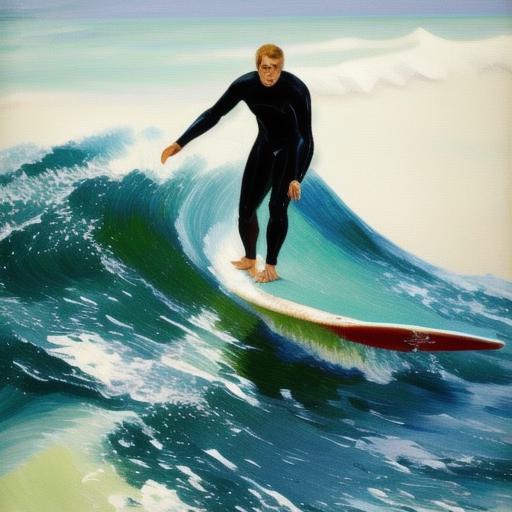}
        \end{subfigure}
        \begin{subfigure}{0.19\textwidth}
            \includegraphics[width=\linewidth]{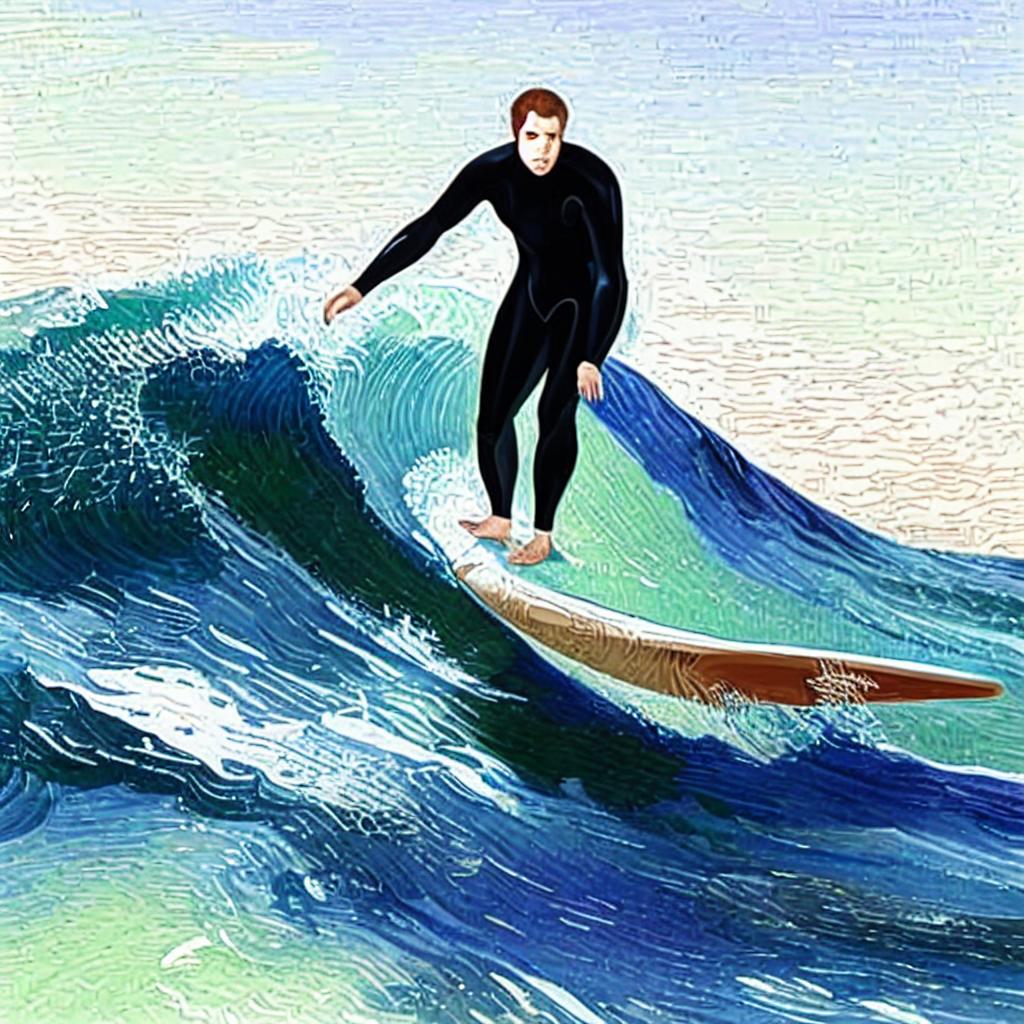}
        \end{subfigure}
        \begin{subfigure}{0.19\textwidth}
            \includegraphics[width=\linewidth]{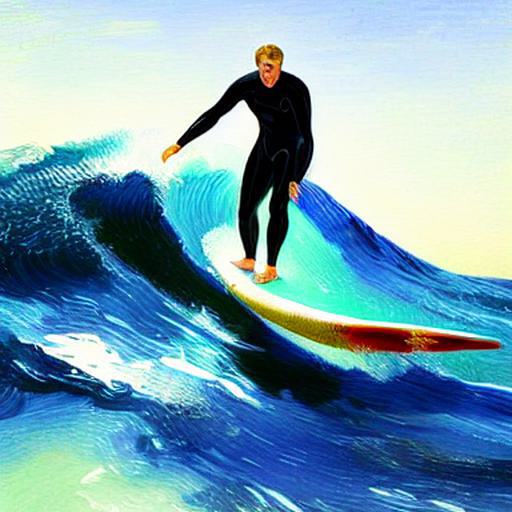}
        \end{subfigure}
            \begin{subfigure}{0.19\textwidth}
            \includegraphics[width=\linewidth]{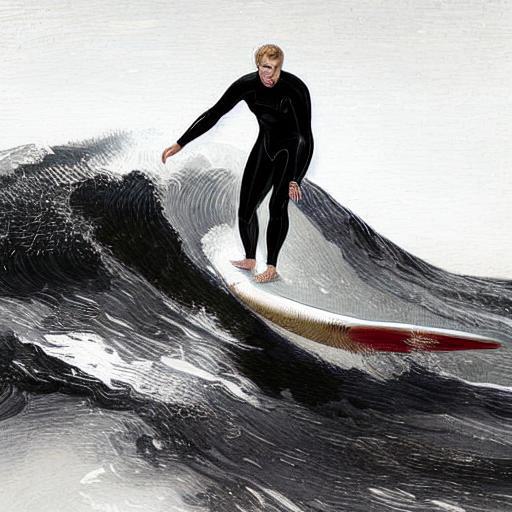}
        \end{subfigure}
    
        \begin{subfigure}{0.19\textwidth}
            \includegraphics[width=\linewidth]{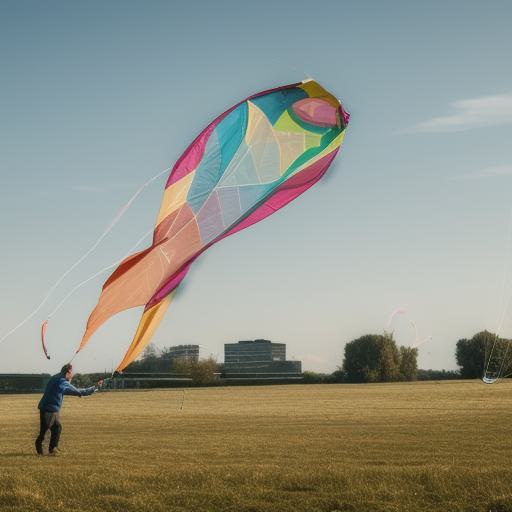}
        \end{subfigure}
        \begin{subfigure}{0.19\textwidth}
            \includegraphics[width=\linewidth]{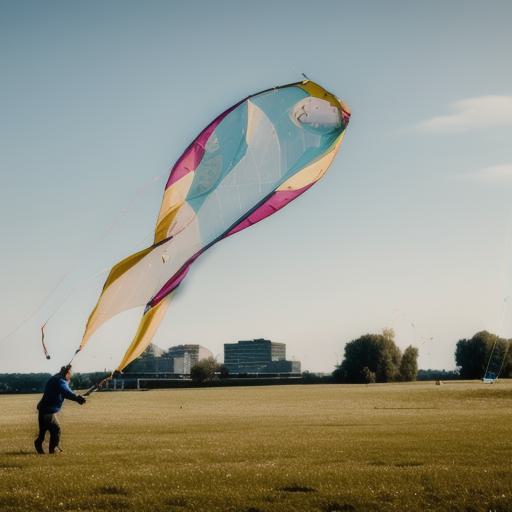}
        \end{subfigure}
        \begin{subfigure}{0.19\textwidth}
            \includegraphics[width=\linewidth]{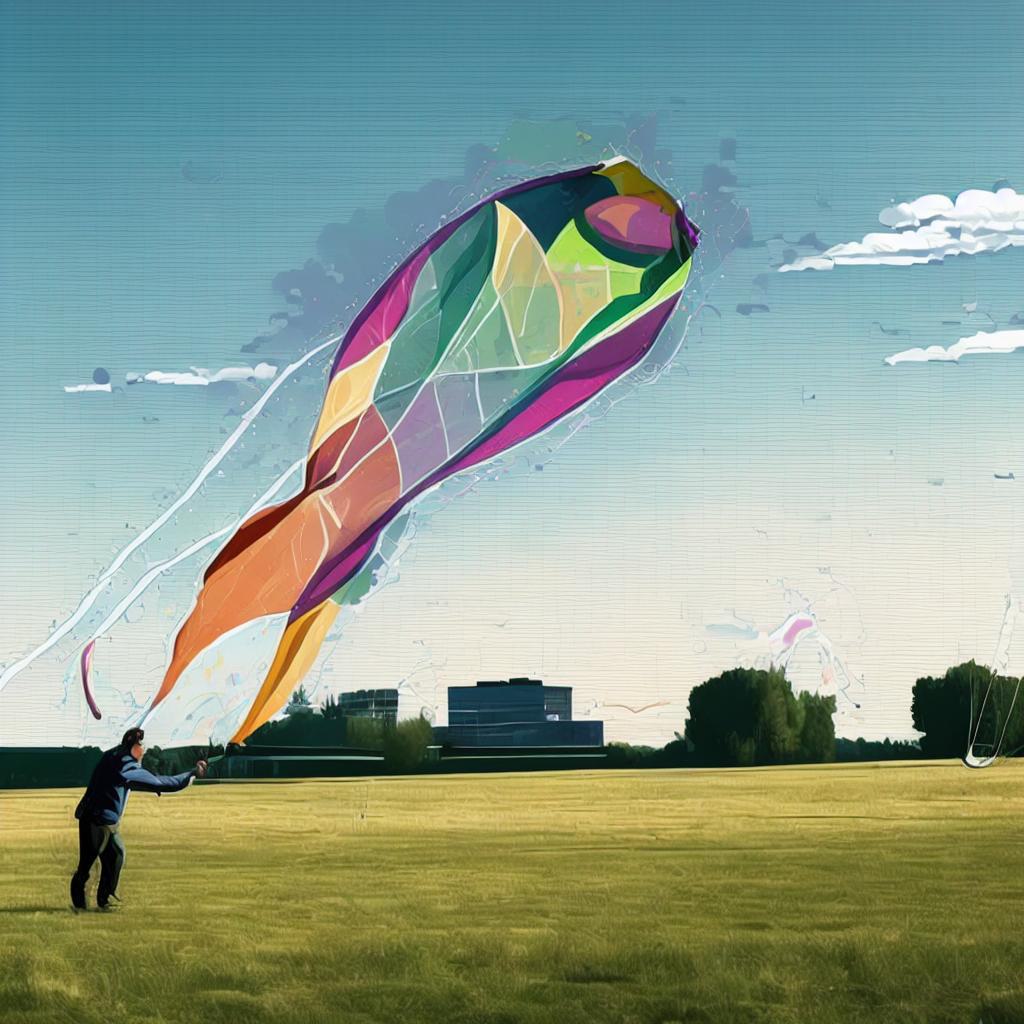}
        \end{subfigure}
        \begin{subfigure}{0.19\textwidth}
            \includegraphics[width=\linewidth]{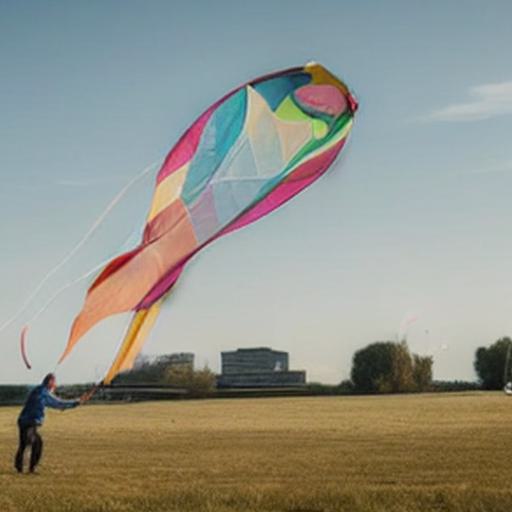}
        \end{subfigure}
            \begin{subfigure}{0.19\textwidth}
            \includegraphics[width=\linewidth]{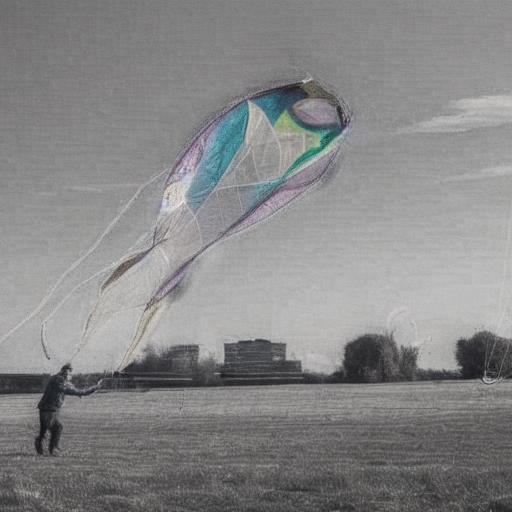}
        \end{subfigure}
    \end{minipage}\\[0.3em]

    \begin{minipage}[c]{0.02\textwidth}
        \rotatebox[origin=c]{90}{Increase depth contrast+highlights}
    \end{minipage}%
    \begin{minipage}[c]{\textwidth}
        \begin{subfigure}{0.19\textwidth}
            \includegraphics[width=\linewidth]{graphics/relatedwork_comp/4200_van_gogh/original/orig.jpg}
        \end{subfigure}
        \begin{subfigure}{0.19\textwidth}
            \includegraphics[width=\linewidth]{graphics/relatedwork_comp/4200_van_gogh/ours/depthPower.jpg}
        \end{subfigure}
        \begin{subfigure}{0.19\textwidth}
            \includegraphics[width=\linewidth]{graphics/relatedwork_comp/4200_van_gogh/stylebooth/depthPower.jpg}
        \end{subfigure}
        \begin{subfigure}{0.19\textwidth}
            \includegraphics[width=\linewidth]{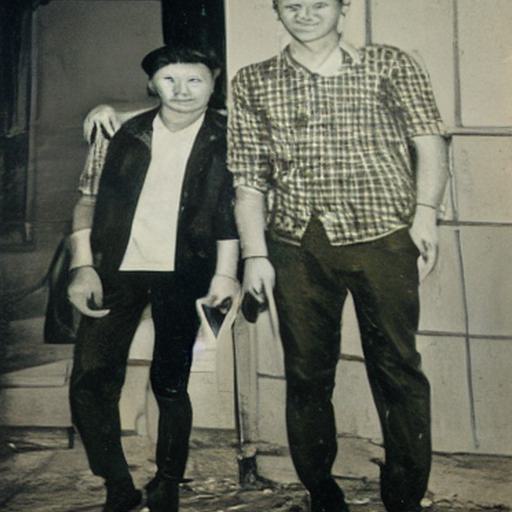}
        \end{subfigure}
            \begin{subfigure}{0.19\textwidth}
            \includegraphics[width=\linewidth]{graphics/relatedwork_comp/4200_van_gogh/ip2p/depthPower.jpg}
        \end{subfigure}
    
        \begin{subfigure}{0.19\textwidth}
            \includegraphics[width=\linewidth]{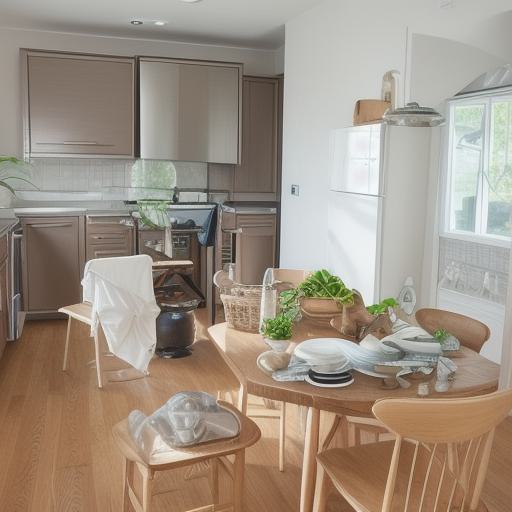}
        \end{subfigure}
        \begin{subfigure}{0.19\textwidth}
            \includegraphics[width=\linewidth]{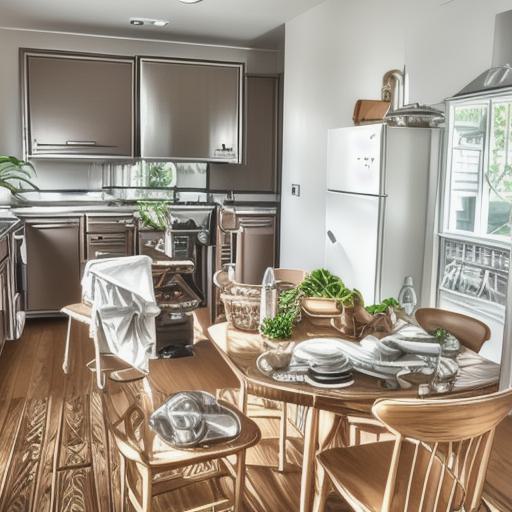}
        \end{subfigure}
        \begin{subfigure}{0.19\textwidth}
            \includegraphics[width=\linewidth]{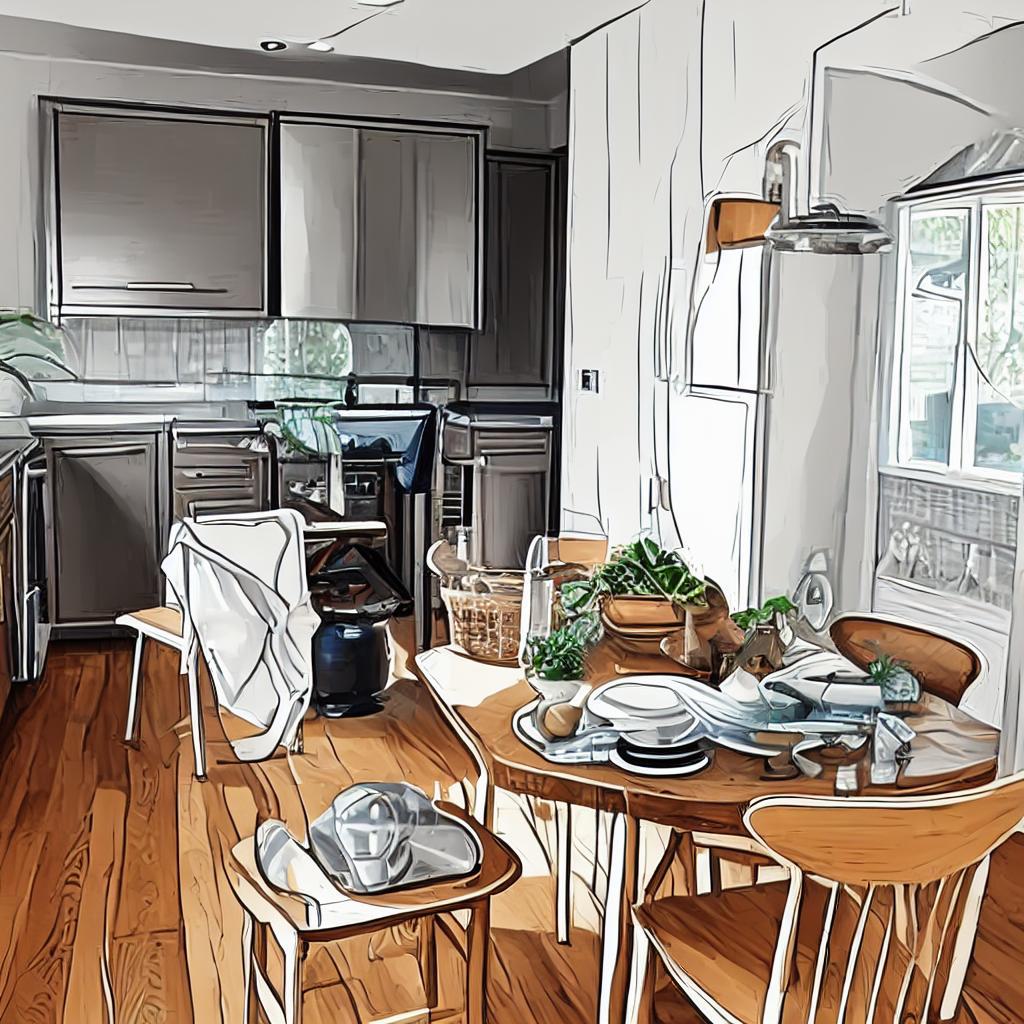}
        \end{subfigure}
        \begin{subfigure}{0.19\textwidth}
            \includegraphics[width=\linewidth]{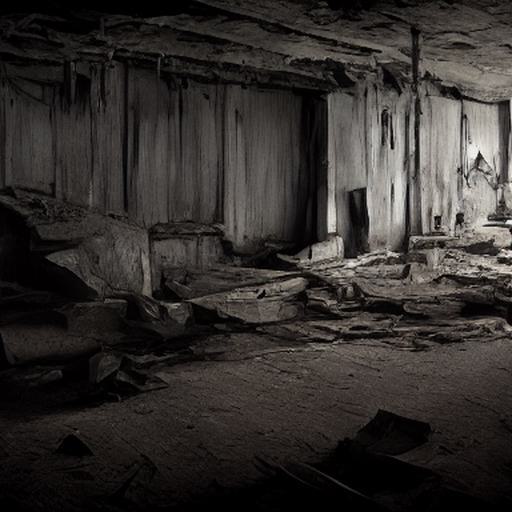}
        \end{subfigure}
            \begin{subfigure}{0.19\textwidth}
            \includegraphics[width=\linewidth]{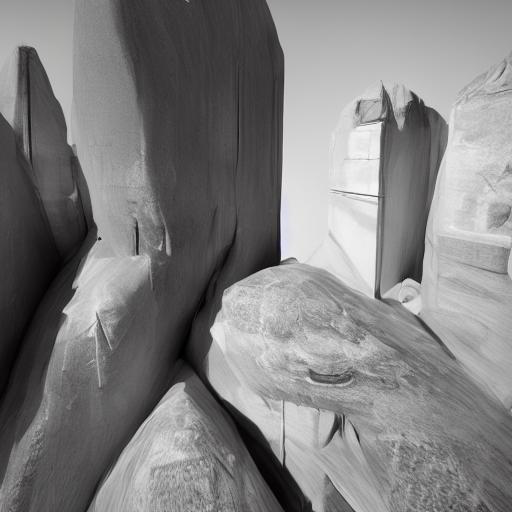}
        \end{subfigure}
    \end{minipage}
    \caption{Qualitative comparison to related methods. We show results from the user study. Note that for fair comparison, all images were generated at the same parameter or cfg setting.}
    \label{fig:sup:comparison_images_rw_userstudy}
\end{figure*}

To apply the prompt-based editing methods to our testing dataset as described in Sec. 5.1, we use the following prompts (for the instruction based methods), and editing captions (shown to users):

\begin{enumerate}[align=left]
    \item[depth power.] Prompt: ``make it have an intense depth contrast and highlights". Editing description: ``Increase the local contrast and highlights"
    \item[contour width.] Prompt: ``make it have thick strokes". Editing description: ``Increase the number and thickness of outlines in the image"
    \item[black point.] Prompt: ``Lower the black point of the whole image". Editing description: ``Lower the black point of the image"
\end{enumerate}
To generate corresponding results with our method, we set $\lambda_A=1$ for each parameter.  In \cref{fig:userstudy_details} we plot response counts for the user study per style and parameter. Our method is preferred by a significant margin against all other methods for all parameters and styles, except for the blackpoint parameter with Van Gogh style. In \cref{fig:sup:comparison_images_rw_userstudy} we show samples from the study and in \cref{fig:user_study_screenshot} we show a screenshot from the web-interface of the user study.
\vspace{-1em}

\section{Additional Training Evaluation}

\begin{figure}[p]
    \centering
    \includegraphics[width=\linewidth]{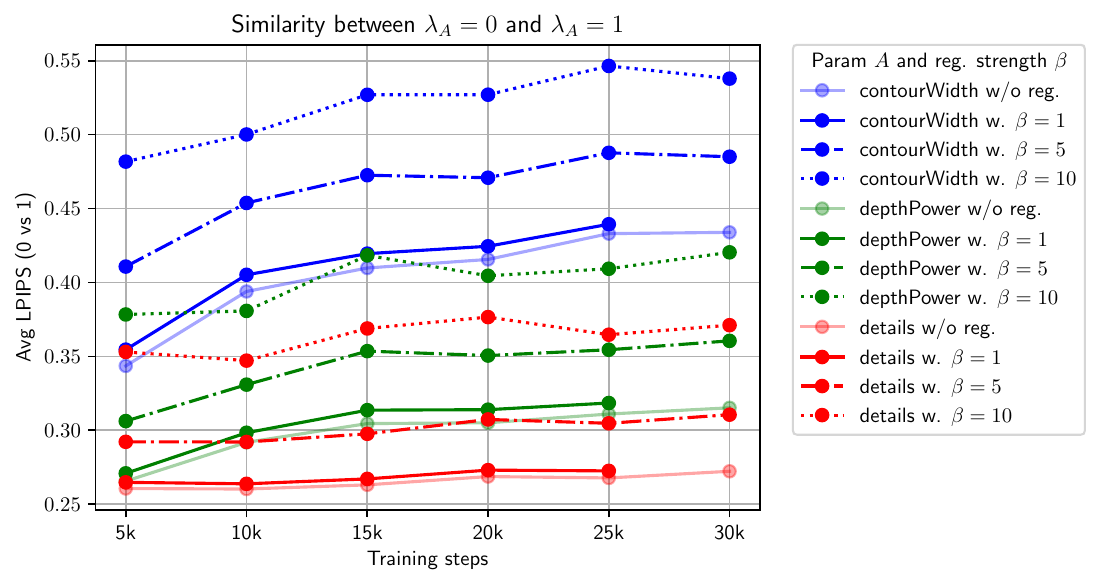}
    \caption{Similarity of parameters over training steps}
    \label{fig:metrics_training_steps}
\end{figure}

We evaluated the impact of training steps and regularization on model performance, as shown in \cref{fig:metrics_training_steps}. To measure visual differences, we used \ac{LPIPS} \cite{zhang2018perceptual} between images generated at $\lambda_A = 0$ and $\lambda_A = 1$ across various parameters. Increasing training steps amplifies the effect of $\lambda$ on outputs, with structural changes like “contourWidth” showing greater visual impact than color or contrast adjustments like “depthPower” or “details.” 

\begin{figure}[h]
    \centering
    \includegraphics[width=\linewidth]{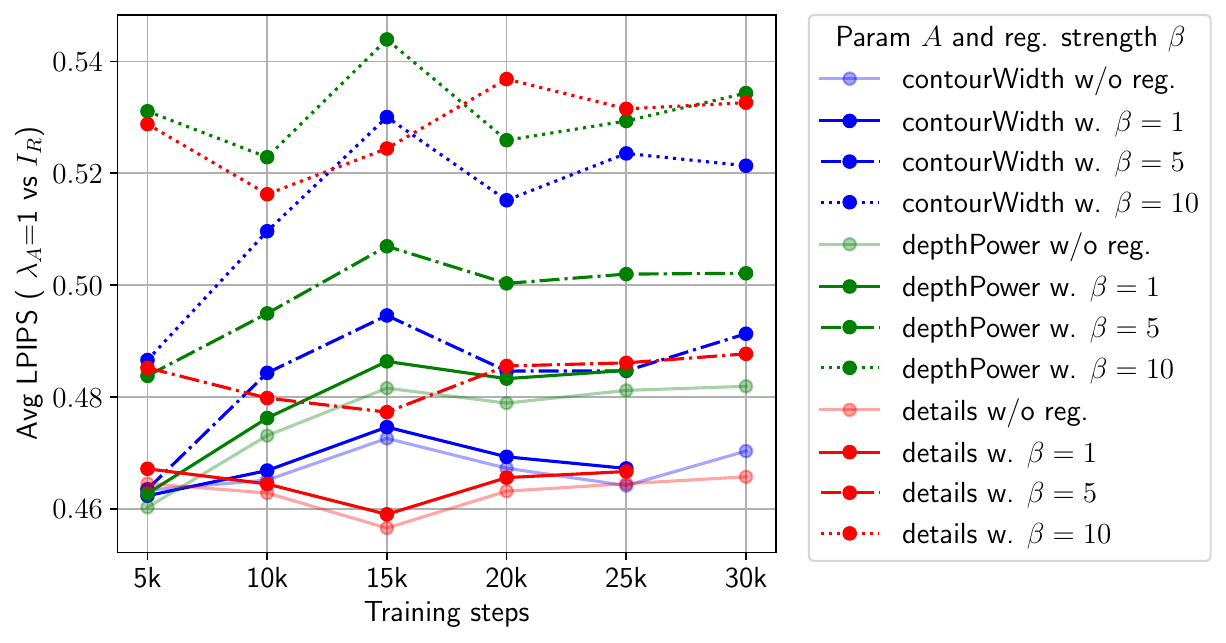}
    \caption{Training behaviour for parameters and regularization strength. We measure the LPIPS between images generated with $\lambda_A=1$ and the original Controlnet outputs $I_R$ filtered using an algorithmic filter, with the same parameter setting. Results are averaged over our testing dataset. Regularization and more steps generally lead to an increase in stylization strength.}
    \label{fig:metrics_training_steps_ir}
\end{figure}

\begin{figure*}[h]
    \centering
    \begin{subfigure}{0.16\linewidth}
        \includegraphics[width=\linewidth]{graphics/cnet_only/lenacanny.jpg}
        \caption{canny edge map}
    \end{subfigure}
    \begin{subfigure}{0.16\linewidth}
        \includegraphics[width=\linewidth]{graphics/cnet_only/blackpoint_egs_7.0_p_1.0_aphotoofawoman.jpg}
        \caption{black point}
    \end{subfigure}
        \begin{subfigure}{0.16\linewidth}
        \includegraphics[width=\linewidth]{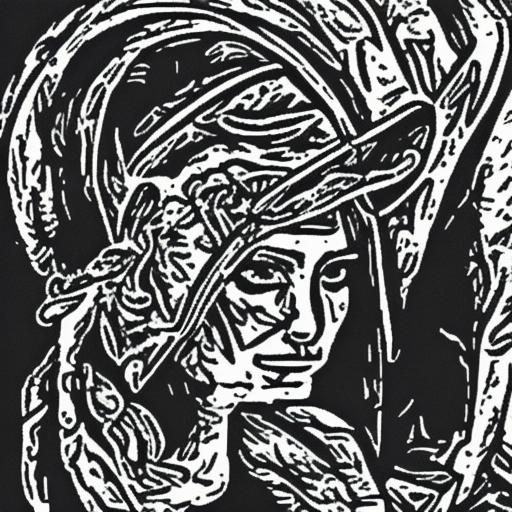}
        \caption{depth power}
    \end{subfigure}
    \begin{subfigure}{0.16\linewidth}
        \includegraphics[width=\linewidth]{graphics/cnet_only/details_egs_7.0_p_1.0_aphotoofawoman.jpg}
        \caption{details}
    \end{subfigure}
    \begin{subfigure}{0.16\linewidth}
        \includegraphics[width=\linewidth]{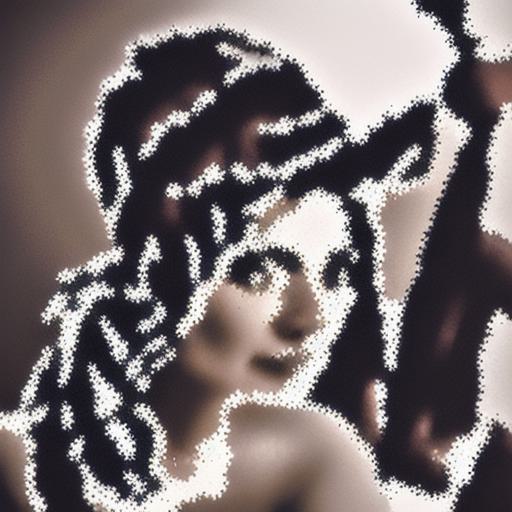}
        \caption{wobbling}
    \end{subfigure}
        \begin{subfigure}{0.16\linewidth}
        \includegraphics[width=\linewidth]{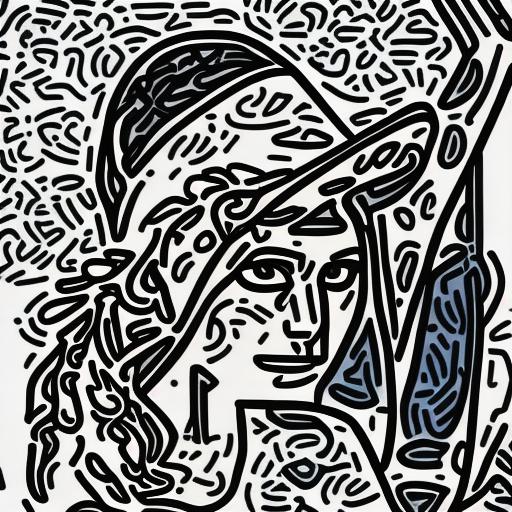}
        \caption{contour width}
    \end{subfigure}
    
    \caption{Extended \Cref{fig:cnet_only_out}. Outputs of only the Controlnet ($g_A$) for different parameters, conditioned on the edge map (a). All parameters are set to strength $\lambda=1$, generating using prompt ``a photo of a woman". }
    \label{fig:sup:cnet_only_out}
\end{figure*}

\begin{figure*}[htbp]
    \centering
    \begin{adjustbox}{valign=m}
        \rotatebox[origin=c]{90}{$\beta=0$}
    \end{adjustbox}
    \begin{adjustbox}{valign=m}
        \begin{subfigure}[b]{0.24\textwidth}
            \includegraphics[width=\textwidth]{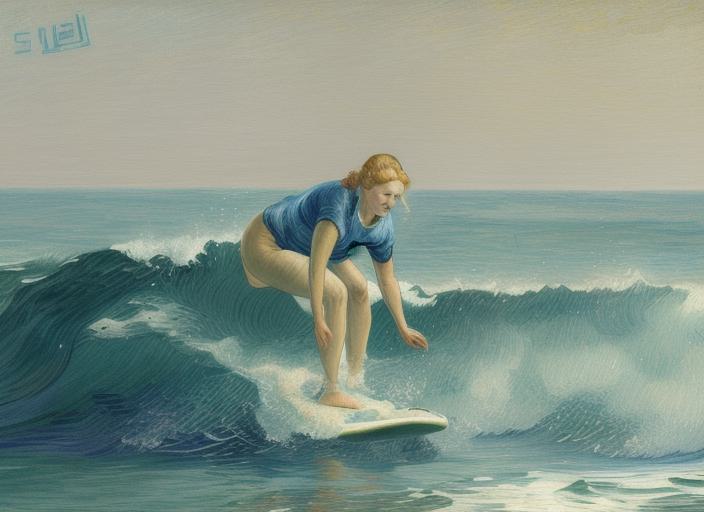}
        \end{subfigure}
        \begin{subfigure}[b]{0.24\textwidth}
            \includegraphics[width=\textwidth]{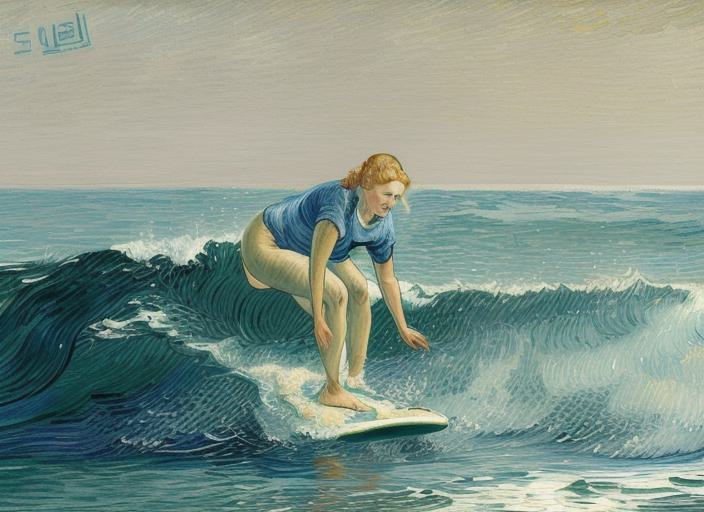}
        \end{subfigure}
        \begin{subfigure}[b]{0.24\textwidth}
            \includegraphics[width=\textwidth]{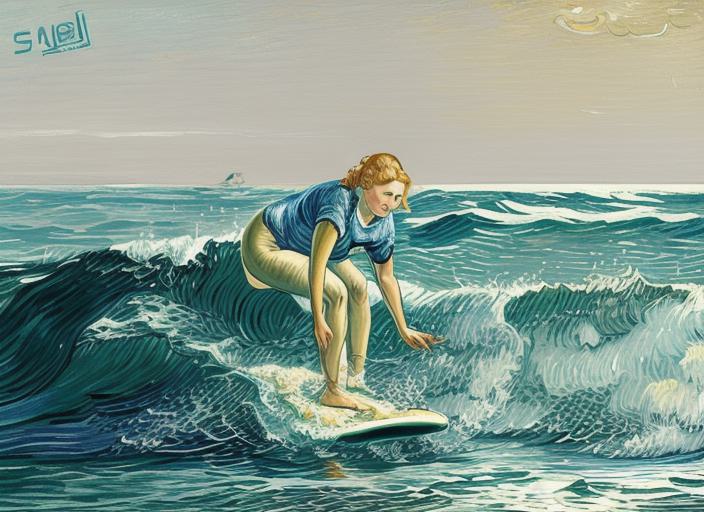}
        \end{subfigure}
        \begin{subfigure}[b]{0.24\textwidth}
            \includegraphics[width=\textwidth]{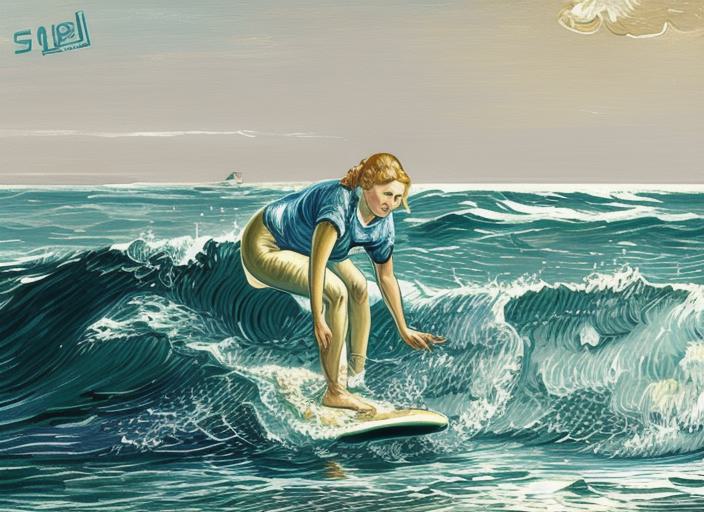}
        \end{subfigure}
        \end{adjustbox}\\[0.1em]

    \begin{adjustbox}{valign=m}
        \rotatebox[origin=c]{90}{\textbf{$\beta=5$}}
    \end{adjustbox}
    \begin{adjustbox}{valign=m}
        \begin{subfigure}[b]{0.24\textwidth}
            \includegraphics[width=\textwidth]{graphics/training_progress/van_gogh_depthpower/orig.jpg}
        \end{subfigure}
        \begin{subfigure}[b]{0.24\textwidth}
            \includegraphics[width=\textwidth]{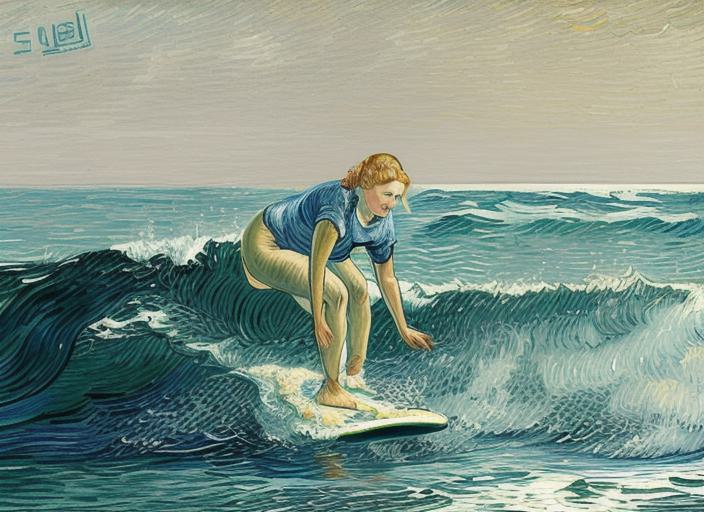}
        \end{subfigure}
        \begin{subfigure}[b]{0.24\textwidth}
            \includegraphics[width=\textwidth]{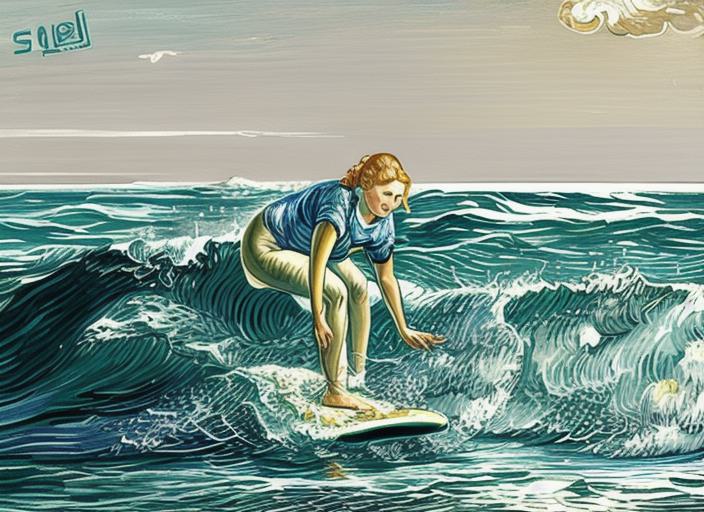}
        \end{subfigure}
        \begin{subfigure}[b]{0.24\textwidth}
            \includegraphics[width=\textwidth]{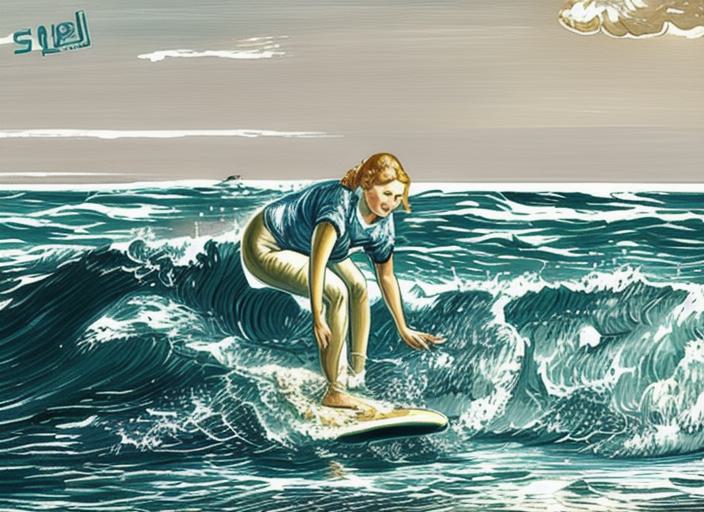}
        \end{subfigure}
    \end{adjustbox}\\[0.1em]

    \begin{adjustbox}{valign=m}
        \rotatebox[origin=c]{90}{\textbf{$\beta=10$}}
    \end{adjustbox}
    \begin{adjustbox}{valign=m}
        \begin{subfigure}[b]{0.24\textwidth}
            \includegraphics[width=\textwidth]{graphics/training_progress/van_gogh_depthpower/orig.jpg}
            \caption*{CNet (orig)}
        \end{subfigure}
        \begin{subfigure}[b]{0.24\textwidth}
            \includegraphics[width=\textwidth]{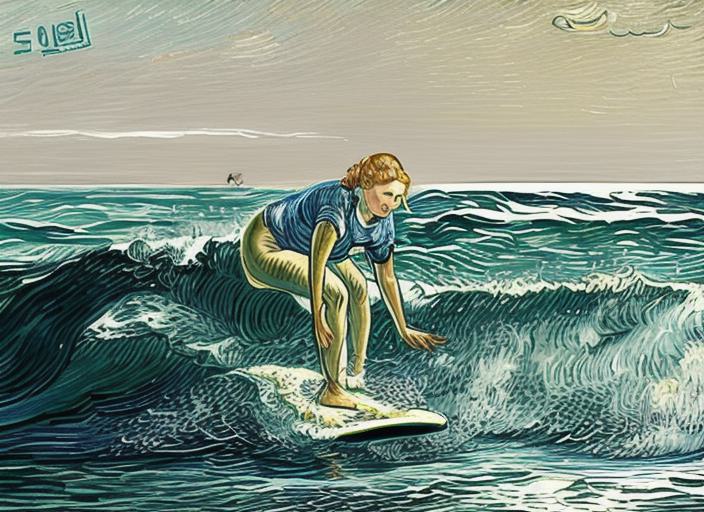}
            \caption*{5k steps}
        \end{subfigure}
        \begin{subfigure}[b]{0.24\textwidth}
            \includegraphics[width=\textwidth]{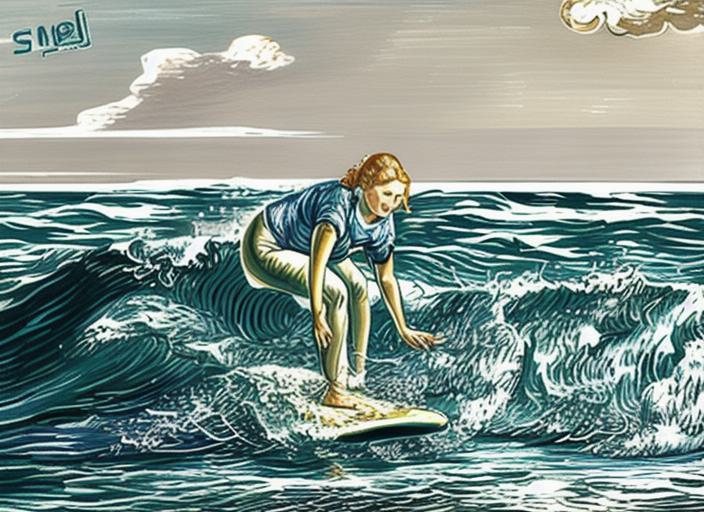}
            \caption*{15k steps}
        \end{subfigure}
        \begin{subfigure}[b]{0.24\textwidth}
            \includegraphics[width=\textwidth]{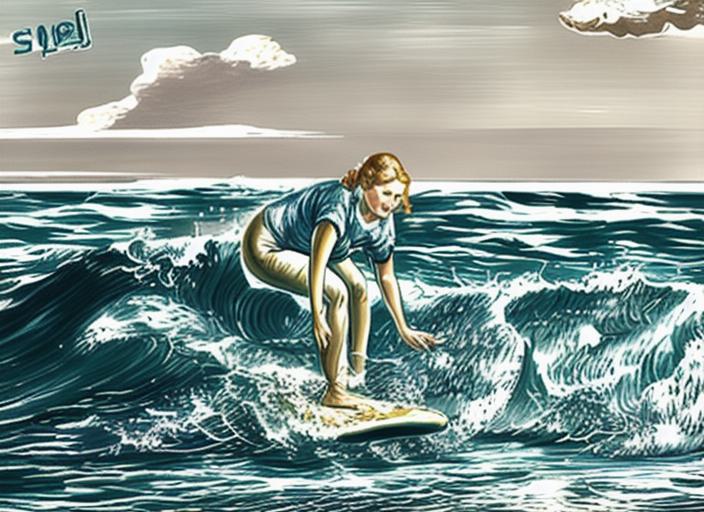}
            \caption*{30k steps}
        \end{subfigure}
    \end{adjustbox}

    \caption{Training progress with different regularizations. Note that the outputs are at $\lambda_{\text{depth power}}=1$, thus at the maximum of the learned range and a strong stylization without significant content alterations from the original image are therefore desireable.}
    \label{fig:sup:training_progress}
\end{figure*}

In \cref{fig:sup:cnet_only_out}, we present additional outputs from the finetuned Controlnet alone (i.e., $\varepsilon_{A,\lambda_A=0} + w_2g_A$), showing that training extra attention layers conditions model outputs on parameter-specific changes. \cref{fig:metrics_training_steps_ir} compares model outputs at $\lambda=1$ with algorithmic filtering on the same Controlnet-generated images $I_R$. Using the same watercolor effect filters as in training, results indicate that higher regularization reduces similarity to algorithmic filtering. However, lower regularization values catch up after about 15k training steps, with similarities converging beyond this point.
In \cref{fig:sup:training_progress}, we illustrate outputs across training steps and regularization settings for the depth power parameter. As with contour width (\Cref{fig:comparison_training_progress}), high regularization quickly reaches strong stylization, but excessive training may intensify stylization and alter content.

\vspace{-1em}
\section{Additional Results}
\begin{figure*}[h]
    \centering
    \begin{subfigure}{0.15\textwidth}
        \includegraphics[width=\linewidth]{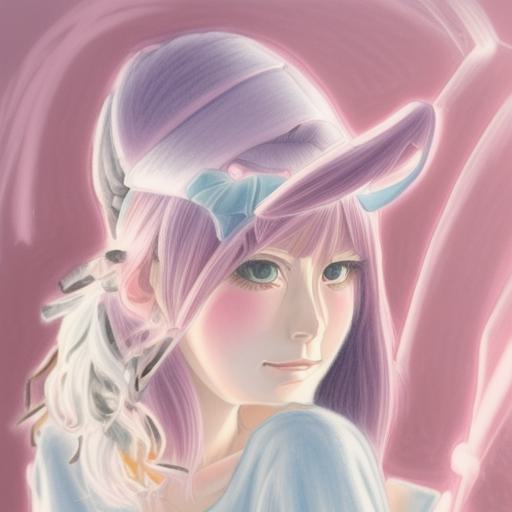}
        \caption{1.2}
    \end{subfigure}
    \begin{subfigure}{0.15\textwidth}
        \includegraphics[width=\linewidth]{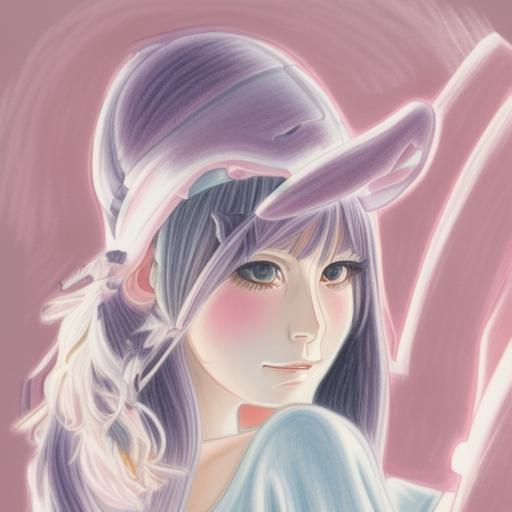}
        \caption{0.95}
    \end{subfigure}
    \begin{subfigure}{0.15\textwidth}
        \includegraphics[width=\linewidth]{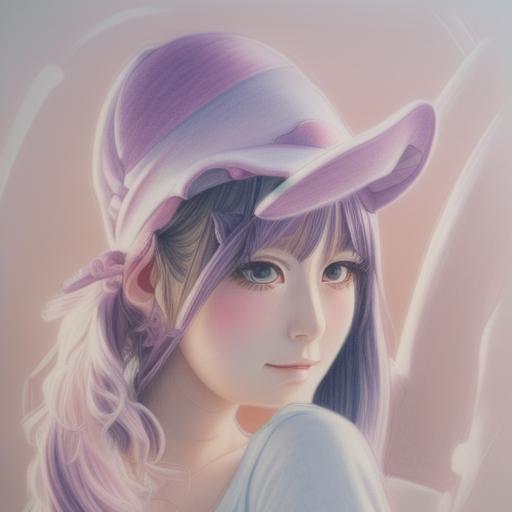}
        \caption{Original}
    \end{subfigure}
    \begin{subfigure}{0.15\textwidth}
        \includegraphics[width=\linewidth]{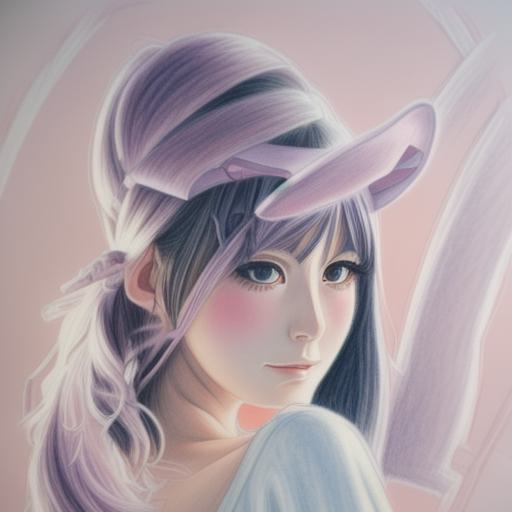}
        \caption{0.65}
    \end{subfigure} 
    \begin{subfigure}{0.15\textwidth}
        \includegraphics[width=\linewidth]{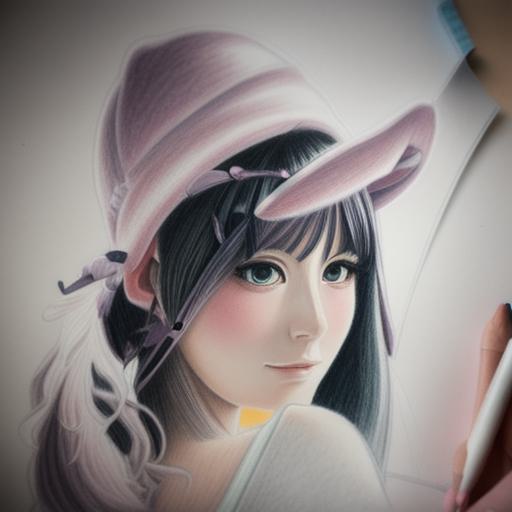}
        \caption{0.35}
    \end{subfigure}
    \begin{subfigure}{0.15\textwidth}
        \includegraphics[width=\linewidth]{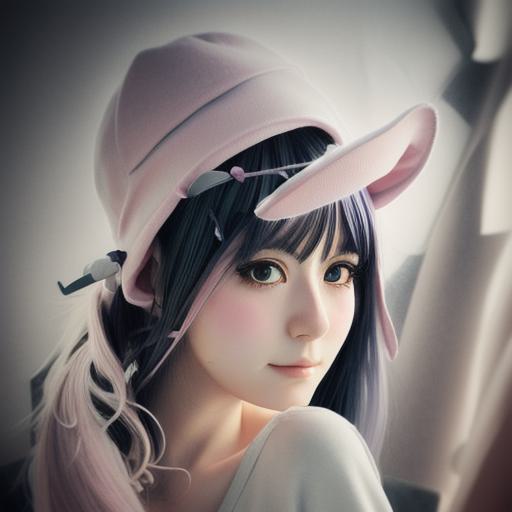}
        \caption{0.1}
    \end{subfigure}
    \caption{Parameter variation for the Artbench-trained model. The parameter encodes the ``expressionism" style, and we use the prompt ``a pastel drawing of a anime style girl" to generate the original. Decreasing the parameter increases the realism of the image since parameter setting 1 encodes an artwork, and 0 a photo. }
    \label{fig:sup:artbench_variation}
\end{figure*}

\begin{figure*}
    \centering
    \includegraphics[width=0.9\linewidth]{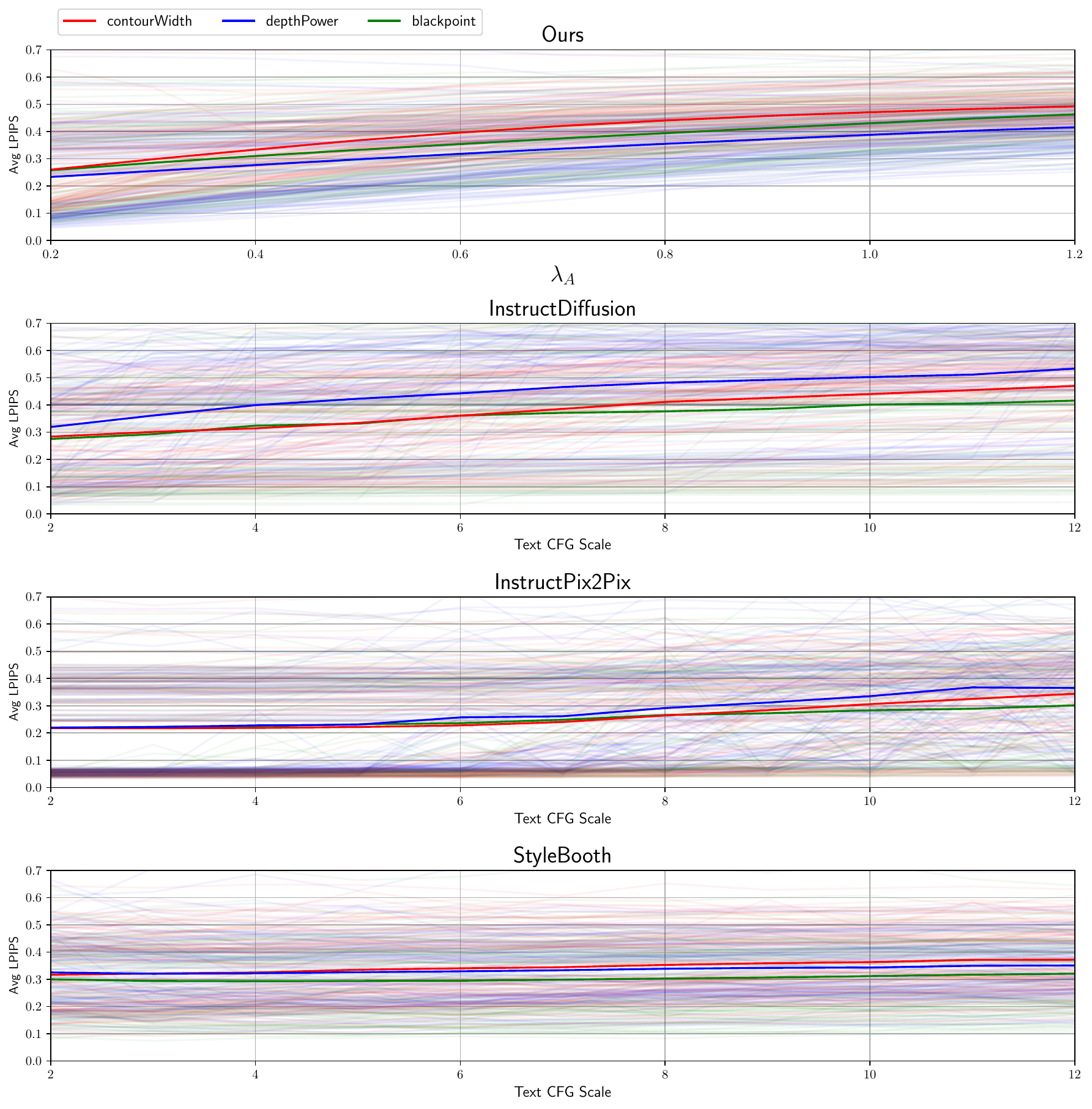}
    \caption{
    Influence of increasing stylization strength - extension of \Cref{fig:parameter_cfg_influence}. For our approach, the parameter ($\lambda_A$) is incrementally adjusted, while text-based methods use classifier-free guidance scale and the editing prompts as defined in \cref{sec:userstudydetails}. Comparisons are made against InstructDiffusion \cite{Geng23instructdiff}, InstructPix2Pix \cite{brooksInstructPix2PixLearningFollow2023}, and StyleBooth \cite{han2024stylebooth}.
    }
    \label{fig:sup:parameter_cfg_influence}
\end{figure*}

\begin{figure*}
    \centering
    \includegraphics[width=1\linewidth]{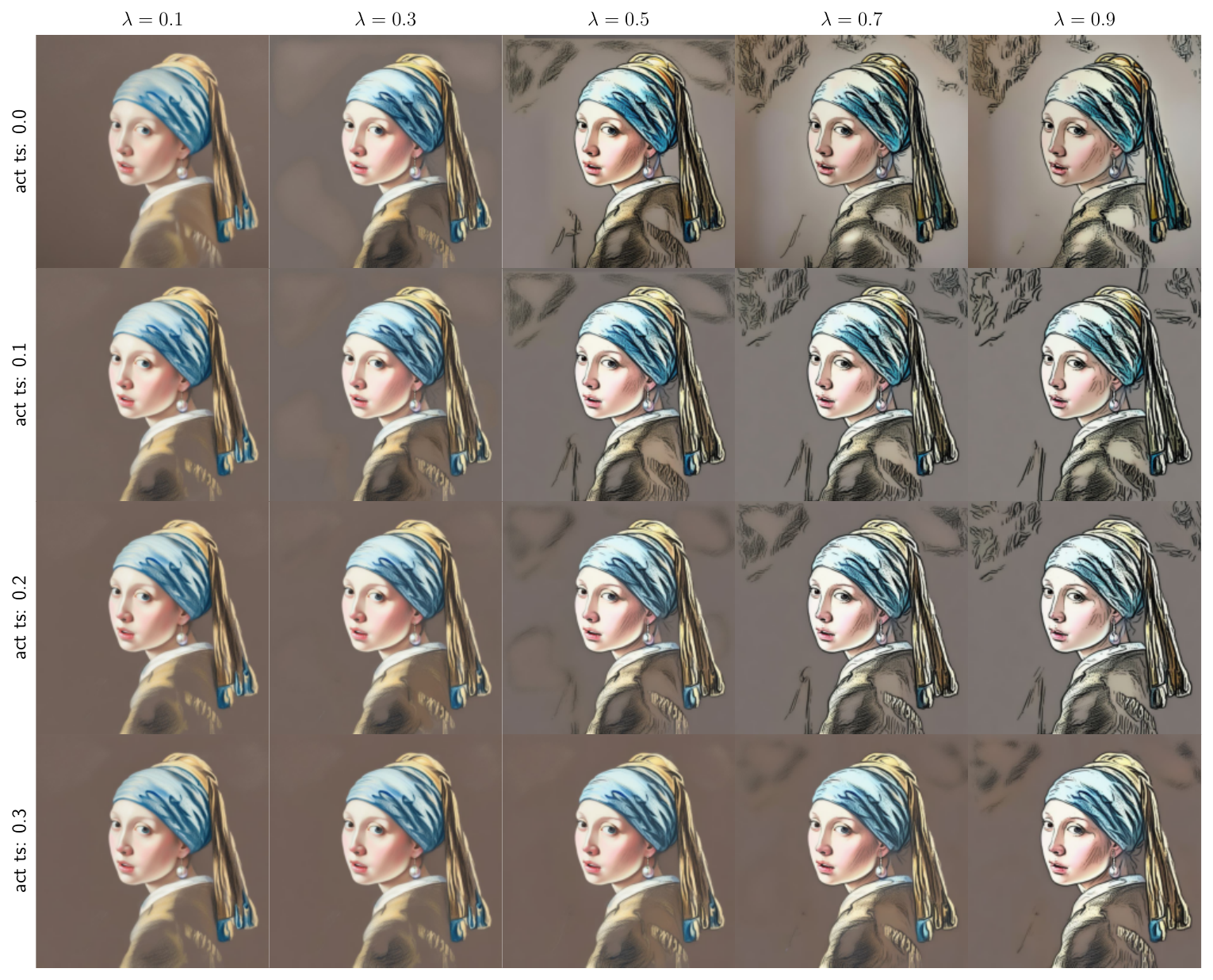}
    \caption{Extended \Cref{fig:vary_act_p}, varying the activation time step and the parameter ($\lambda$). Generation prompt: ``a pastel drawing of a girl with a pearl earring".}
    \label{fig:sup:vary_act_p}
\end{figure*}

In \cref{fig:sup:artbench_variation}, we show results for the artbench-finetuned model.
In \cref{fig:sup:vary_act_p}, we vary both the activation timestep and parameter to show their respective influences. A higher (=later) activation step introduces less changes into the image, representing a tradeoff between stability of the image and strenghth of stylization. In \cref{fig:sup:parameter_cfg_influence}, we plot the stylization stability ($\lambda$ vs LPIPS) for all methods from our quantitative comparison. Notably, some text-based approaches exhibit minimal LPIPS change, indicating  either limited sensitivity to text guidance scale or ineffective stylization even with increased guidance scale. Others display abrupt jumps in LPIPS, making fine-grained stylization adjustments infeasible.

